\documentclass[11pt]{article}

\usepackage[utf8]{inputenc}
\usepackage[T1]{fontenc}
\usepackage{fullpage}
\usepackage{url}
\usepackage{booktabs}
\usepackage{amsfonts}
\usepackage{nicefrac}
\usepackage{microtype}
\usepackage{xcolor}
\usepackage{colortbl}
\usepackage{graphicx}
\usepackage{mathtools}
\usepackage{amsmath}
\usepackage{amssymb}
\usepackage{amsthm}
\usepackage{mathbbol}
\usepackage{mathrsfs}
\usepackage[most]{tcolorbox}
\usepackage[numbers,square]{natbib}

\usepackage{subcaption}
\usepackage{hyperref}
\usepackage[nameinlink]{cleveref}
\crefname{subtable}{table}{tables}
\Crefname{subtable}{Table}{Tables}
\usepackage{placeins}
\usepackage{needspace}
\DeclareRobustCommand{\prism}{\ensuremath{\mathtt{PRISM}}}
\DeclareRobustCommand{\alphaprism}{\ensuremath{\alpha\text{-}\prism{}}}
\newcommand{\yhat}{\widehat{y}}
\hypersetup{
  colorlinks,
  hypertexnames=false,
  linkcolor={red!72!black},
  citecolor={blue!72!black},
  urlcolor={magenta!72!black},
  pdftitle={PRISM: Calibrating Black-Box Judges with Counterfactual Probes},
  pdfauthor={Kakei Yamamoto; So Kuroki}
}

\makeatletter
\long\def\@makecaption#1#2{\vskip\abovecaptionskip
  \small #1: #2\par
  \vskip\belowcaptionskip
}
\renewcommand{\section}{\@startsection{section}{1}{\z@}{-1.85ex \@plus -0.45ex \@minus -0.2ex}{1.25ex \@plus 0.25ex \@minus 0.2ex}{\large\bfseries\raggedright}}
\renewcommand{\subsection}{\@startsection{subsection}{2}{\z@}{-1.6ex \@plus -0.45ex \@minus -0.2ex}{0.65ex \@plus 0.2ex}{\normalsize\bfseries\raggedright}}
\renewcommand{\paragraph}{\@startsection{paragraph}{4}{\z@}{1.1ex \@plus 0.4ex \@minus 0.2ex}{-1em}{\normalsize\bfseries}}
\newcommand{\blfootnote}[1]{\begingroup
  \renewcommand{\thefootnote}{}\footnote{#1}\addtocounter{footnote}{-1}\endgroup
}

\makeatother

\AtBeginDocument{\setlength{\parskip}{4.5pt plus 0.5pt minus 0.5pt}
  \setlength{\textfloatsep}{8pt plus 1pt minus 2pt}
  \setlength{\floatsep}{7pt plus 1pt minus 2pt}
  \setlength{\intextsep}{7pt plus 1pt minus 2pt}
  \setlength{\abovecaptionskip}{3.5pt plus 1pt minus 1pt}
  \setlength{\belowcaptionskip}{0pt}
  \setlength{\abovedisplayskip}{5.5pt plus 1.5pt minus 1.5pt}
  \setlength{\belowdisplayskip}{5.5pt plus 1.5pt minus 1.5pt}
  \setlength{\abovedisplayshortskip}{3.5pt plus 1pt minus 1pt}
  \setlength{\belowdisplayshortskip}{3.5pt plus 1pt minus 1pt}
}

\newcommand{\myparagraph}[1]{\noindent{\bfseries #1:}}
\newcommand{\mygraybox}[1]{\begin{tcolorbox}[
    colback=gray!10, colframe=gray!70, arc=4pt, boxrule=0.7pt,
    left=7pt, right=7pt, top=4pt, bottom=4pt,
    before skip=6pt plus 2pt minus 2pt,
    after skip=6pt plus 2pt minus 2pt,
    enhanced jigsaw, parbox=false]
    #1
  \end{tcolorbox}}
\newcommand{\real}{\mathbb{R}}
\newcommand{\Exs}{\mathbb{E}}
\newcommand{\defn}{\coloneqq}
\newcommand{\etalchar}[1]{\ensuremath{^{#1}}}
\DeclareMathOperator*{\argmin}{arg\,min}

\begin{document}
\begin{center}
{\bfseries\LARGE \prism{}: Calibrating Black-Box Judges with Counterfactual Probes\par}
\vspace*{0.2in}
  \begin{tabular}{ccc}
    Kakei Yamamoto$^\dagger$ && So Kuroki$^\star$\\
    \texttt{kakei@mit.edu} && \texttt{sokuroki@sakana.ai}
  \end{tabular}

  \vspace*{0.2in}
  \begin{tabular}{cc}
    Lab for Information and Decision Systems$^{\dagger}$ & Sakana AI$^{\star}$\\
    Statistics and Data Science Center & \\
    EECS & \\
    Massachusetts Institute of Technology &
  \end{tabular}\par

\vspace*{0.25in}
October 13, 2026
\vspace*{0.25in}
\begin{abstract}
Large language models are increasingly used as judges to reduce
reliance on costly human evaluation, yet their scores can
systematically disagree with human judgments. We ask whether responses
to an explicitly imposed evaluative bias can help correct a judge's
implicit scoring biases. We introduce \prism{} (Probe-based
Residual Inference of Scoring Misalignment), which holds the evaluated
content fixed and measures score changes when attributes are emphasized
or de-emphasized. A local sensitivity model motivates representing
these responses as a surrogate Jacobian. Ridge regression on human
labels learns a correction to the baseline score. The resulting
scoring protocol is rubric-free: it uses overall scores without
requesting criterion-level ratings. In the main Jev and Sol comparisons on
SummEval summarization and HANNA story evaluation, \prism{} and its
learned-baseline variant significantly improve agreement with human
rankings over uncalibrated scores while remaining competitive with
rubric-based methods. A controlled SummEval comparison yields similar
agreement for rubric-derived quality probes and style probes,
supporting calibration without reproducing rubric criteria. The learned
coefficients reveal shared and opposing correction directions across
judges.
\end{abstract}
\end{center}
\blfootnote{This work was carried out while Kakei Yamamoto was an intern at Sakana AI.}

\section{Introduction}
\label{SecIntroduction}

Human evaluation provides a reference for assessing generated text
\citep{fabbri2021summeval}. Large language models (LLMs) offer scalable
evaluation \citep{zheng2023judging} and a source of training feedback
that reduces reliance on costly human labels \citep{lee2024rlaif}.
Recent releases such as Jev extend this activity to decision models
with typed probabilistic outputs for scoring and verification
\citep{almeida2026jev,deusser2026jev}. Yet LLM judgments can favor verbosity,
particular answer positions, or their own generations
\citep{zheng2023judging,wang2024fair,shi2025judges,panickssery2024selfpreference}. On subjective
evaluation tasks, judges can also agree with one another while
diverging from human judgments \citep{mukherjee2026geometry}.
Improving agreement with human judgments is therefore important both for
evaluation and for the feedback supplied to learning systems.

A common approach makes evaluation criteria explicit through rubrics
\citep{hashemi2024rubric,rao2026autorubric,gupta2025carmo,liu2026openrubrics}. G-Eval elicits
ratings using criterion definitions and generated evaluation steps
\citep{liu2023geval}. Recent work also examines the reliability of
LLM-generated rubrics \citep{siro2026learning,zhou2026rubricbench}. SAJA collects a vector
of rubric assessments in one call and fits a supervised prediction
head to human labels \citep{kola2026saja}. Its features describe
criterion-level judgments. We study a complementary observable: how
the overall score changes when evaluative emphasis shifts.

A judge may implicitly overemphasize or underemphasize attributes
relative to human evaluators. Our intuition is to probe these tendencies
by imposing an explicit evaluative bias: favor or discount an attribute
while holding the content fixed, and measure the score change. The
response measures sensitivity to the intervention. These \emph{probes}
draw on perturbation analysis, which studies model behavior through
controlled input changes \citep{ribeiro2016trust,cohenwang2024contextcite}.
We ask:

\begingroup
\setlength{\leftmargini}{1em}
\begin{quote}
\raggedright
\centering
\textit{Can perturbation responses guide the debiasing of LLM judges?}
\end{quote}
\endgroup

\begin{figure}[!htbp]
  \centering
  \includegraphics[width=\linewidth]{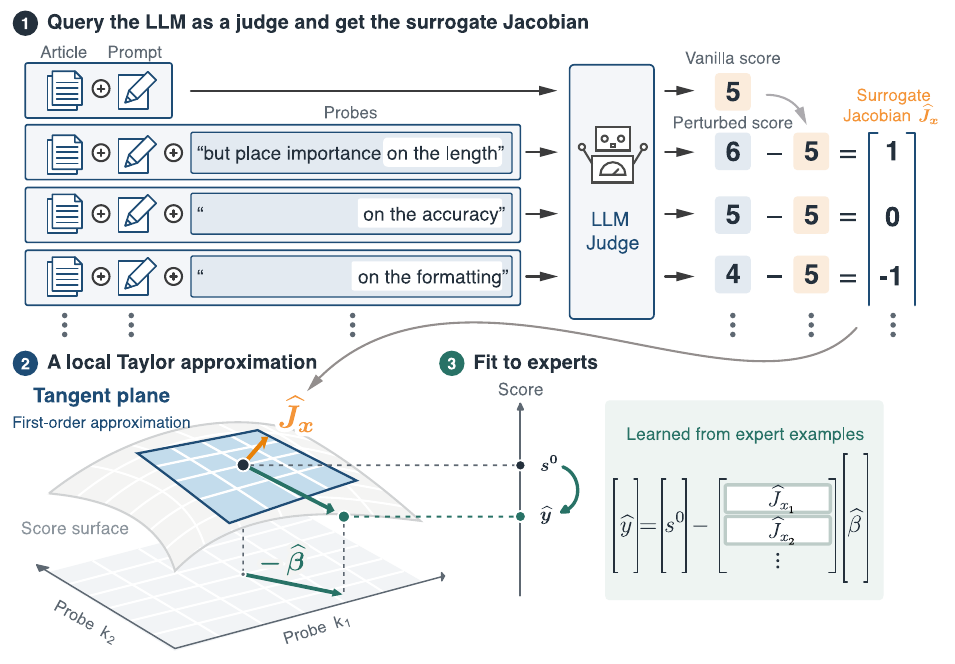}
  \caption{\prism{} overview. (1) Holding the evaluated content fixed,
we ask the judge to change its emphasis on individual attributes.
Differences from the Vanilla score form the surrogate Jacobian
$\widehat J_x$: the illustrated scores $6,5,4$ give responses
$+1,0,-1$ relative to the baseline $5$. Only increased-emphasis probes
are pictured; the method retains both increased- and decreased-emphasis
responses. (2) The tangent plane illustrates the local sensitivity
interpretation: a small change in evaluative weights produces an
approximately linear score change. (3) Ridge regression on human-labeled
examples learns shared correction coefficients $\widehat\beta$; new
inputs are scored as $\yhat=s^0-\widehat J_x\widehat\beta$ using only
baseline and probed judge outputs. The scores, surface, and displacement
$-\widehat\beta$ are schematic, rather than recovered internal model
quantities.}
  \label{FigMethodOverview}
\end{figure}

We introduce \prism{} (Probe-based Residual Inference of Scoring
Misalignment), which uses these responses to learn a correction
toward human judgments.
\Cref{FigMethodOverview} illustrates the method. A local sensitivity
model motivates arranging the measured score changes as a
\emph{surrogate Jacobian} (\cref{SecFormulation}). Ridge regression on
human labels fits the judge's discrepancy from human judgments as a
combination of these responses; subtracting it corrects the baseline
score. This yields a rubric-free scoring protocol: inference requires
only baseline and probed overall scores, without criterion-level ratings.

Our main comparisons cover SummEval summarization
\citep{fabbri2021summeval} and HANNA story evaluation
\citep{chhun2022hanna}. For Jev and Sol, \prism{} and its
learned-baseline variant \alphaprism{} significantly improve ranking
agreement over the uncalibrated judge (Vanilla), while remaining
competitive with rubric-based SAJA and G-Eval. We also examine how
the fitted correction uses probe responses and how calibration depends
on the choice of probes and regression design.

\myparagraph{Organization}
\Cref{SecFormulation} establishes the method.
\Cref{SecCalibration} defines the probe-response representation and
supervised residual correction; \cref{SecInterpretation} tests whether
the responses provide a reproducible signal beyond repeated-score
variation and how they depend on presentation order and evaluative
emphasis. \Cref{SecResults} evaluates the resulting calibration.
\Cref{SecExpertAgreement} compares human-ranking agreement with
Vanilla and rubric-based baselines on summarization and story evaluation.
\Cref{SecCorrectionProfiles} decomposes the fitted correction into
attribute-level contributions to identify shared and differing
correction directions across judges. \Cref{SecAblations} examines
the roles of quality versus style attributes, probe selection,
regularization, and learning the baseline coefficient.
The appendices provide protocols, additional diagnostics, and full
four-judge results.

\subsection{Related work}
\label{SecRelated}

\myparagraph{Evaluator design and calibration}
Specialized evaluators such as Prometheus~2 are trained for direct
assessment and pairwise ranking under user-defined criteria
\citep{kim2024prometheus2}. SAJA instead calibrates a fixed judge and
also analyzes the importance of its rubric features
\citep{kola2026saja}. Recent work fits increasing affine maps to
human-rated anchors, aligning score scales while preserving rankings
\citep{zufle2026calibrating}.
Our features measure how overall scores change under shifted emphasis,
rather than the level of each criterion rating. The fitted coefficients
describe corrections in response coordinates, without identifying
the judge's internal weights.

\myparagraph{Bias diagnosis and correction}
Position bias varies across judges and tasks \citep{shi2025judges}.
CALM diagnoses biases by modifying answers or instructions and
measuring changes in scores or preferences \citep{ye2025justice}.
CalibraEval learns order-preserving maps of selection probabilities
from consistency under option-position and identifier swaps, without
human labels \citep{li2025calibraeval}. \prism{} learns a correction
from human labels and allows the probes to change the judgment;
their predictive value is tested against held-out human evaluations.

\myparagraph{Input perturbations and attribution}
Input leave-one-out attribution measures response changes after
removing individual input parts. LIME fits local surrogates to
perturbed inputs \citep{ribeiro2016trust}; ContextCite fits a sparse
linear surrogate to transformed response probabilities under context
ablations \citep{cohenwang2024contextcite}. Here, the evaluated content
stays fixed and the measured responses are used for score correction.
Influence functions instead approximate the effects of training-example
reweighting or removal \citep{koh2017influence}.

\section{Problem and method formulation}
\label{SecFormulation}

Let $X$ contain the evaluated output and context, and let
$Y\in\real$ be a human reference score for a fixed evaluation target. In a joint
judgment, $X$ also identifies the candidate within the presented set.
The target is $\Exs[Y\mid X=x]$ for the specified rater population.

For a fixed judge, $s(x;p)$ is the numerical score under prompt $p$,
or its mean over experiment-specified repetitions and wordings.
The prompt fixes the task, instructions, scale, and response format.
Write $s^0(x)\defn s(x;p_0)$ for the baseline prompt $p_0$;
\emph{Vanilla} uses this score without a learned correction.
Human and judge scores must be placed on comparable scales before fitting.

\subsection{Probe responses and residual calibration}
\label{SecCalibration}

\myparagraph{Probe responses} For attribute $k$, the prompt $p_k^+$
adds an instruction to give that attribute greater weight, and $p_k^-$
adds an instruction to give it less weight. The evaluated content,
scoring task, output format, and score scale stay fixed. The signs
refer to emphasis, so a positive probe can lower a candidate's score. Define the two oriented responses by
\begin{align}
  \Delta_k^+(x)&\defn s(x;p_k^+)-s^0(x), \notag\\
  \Delta_k^-(x)&\defn s^0(x)-s(x;p_k^-).
  \label{EqnResponse}
\end{align}
We retain these as separate columns of the observed response row,
which we call the \emph{surrogate Jacobian}:
\[
  \widehat J(x)=(\Delta_1^+,\Delta_1^-,\ldots,\Delta_K^+,\Delta_K^-).
\]
The row $\widehat J(x)\in\real^{1\times 2K}$ represents instruction
sensitivity for the linear correction below.
The paired half-difference is a lower-dimensional alternative:
$[\widehat J^{\mathrm{half}}(x)]_k
=(\Delta_k^+(x)+\Delta_k^-(x))/2$.
It combines the two responses into one column per attribute.
Keeping both columns also retains the second difference
\begin{align*}
  \Delta_k^+(x)-\Delta_k^-(x)
  &=s(x;p_k^+)+s(x;p_k^-)\\
  &\quad-2s^0(x),
\end{align*}
which the half-difference alone discards. Thus, \emph{surrogate
Jacobian} denotes an empirical sensitivity representation that
also retains second-difference information, rather than a pure
first-order derivative estimate. No numerical intervention strength
is assigned to the probe wording.

\myparagraph{Example}
\Cref{TabPromptExample} shows a recorded coverage probe: emphasizing
coverage lowers the score, whereas de-emphasizing it raises the score,
with the article and summary held fixed. Additional attribute probes
for the same input appear in \cref{AppPromptExample}.

\begin{table}[!htbp]
\mygraybox{
\small
\noindent\textbf{Example: a coverage probe.}
The following are recorded GPT-5.6-sol responses for one SummEval
candidate. All three calls received the same article and summary;
only the added system instruction changed. The article is excerpted
here for space.\par
\noindent\textbf{System.} You are evaluating machine-generated summaries
of news articles.\par
\noindent\textbf{Source article.} Barcelona club president Josep Maria
Bartomeu has insisted that the La Liga leaders have no plans to replace
Luis Enrique and they're `very happy' with him. $\ldots$\par
\noindent\textbf{Summary.} Barca have no plans to replace Luis Enrique\par
\noindent\textbf{Task.} Rate the overall quality of the summary as a
summary of the article---how well it selects what matters, holds
together as a text, and would serve a reader in place of the article.
Use the whole 1--10 scale: 1--2 = very poor, 3--4 = poor,
5--6 = acceptable, 7--8 = good, 9--10 = excellent.
Respond with only a JSON object: \texttt{\{"score": <int 1--10>\}}.\par
\smallskip
\noindent\begin{tabular}{@{}p{0.09\linewidth}p{0.61\linewidth}p{0.22\linewidth}@{}}
\toprule
Prompt & Added system instruction & Returned output\\
\midrule
$p_0$ & None & \texttt{\{"score": 6\}}\\[3pt]
$p_k^+$ & When you assign the score, weight how much of the article's
important content the summary includes more heavily than you normally
would. & \texttt{\{"score": 5\}}\\[3pt]
$p_k^-$ & When you assign the score, weight how much of the article's
important content the summary includes less heavily than you normally
would. & \texttt{\{"score": 7\}}\\
\bottomrule
\end{tabular}\par
\smallskip
\noindent These calls give $\Delta_k^+=5-6=-1$ and
$\Delta_k^-=6-7=-1$, hence a paired half-difference of $-1$.
}
\caption{Recorded coverage-probe example. The article is excerpted, but
all three calls received the same full article and candidate summary.}
\label{TabPromptExample}
\end{table}

The probe bank fixes the attributes, exact instructions, and score
contrasts. Multiple wordings can be averaged within each sign before
forming the contrasts. A context probe, such as asking the judge to
score for a particular audience, need not have a positive--negative
pair; its column is simply $s(x;p)-s^0(x)$. We write $q$ for the
number of columns in the chosen bank. Different columns may be
correlated. Probe construction uses the calibration portion of the
data; evaluation documents are excluded from fitting the correction.

\myparagraph{Residual calibration} Given calibration data
$\mathcal D_n=\{(x_i,y_i)\}_{i=1}^n$, we fit one coefficient vector
$\beta\in\real^q$ for the judge by ridge regression with
regularization parameter $\lambda>0$:
\begin{equation}
  \widehat\beta\defn\argmin_{\beta\in\real^q}\Bigg\{
  \lambda\|\beta\|_2^2
  +\frac{1}{n}\sum_{i=1}^n
  \big(s^0(x_i)-y_i-\widehat J(x_i)\beta\big)^2\Bigg\}.
\label{EqnRidge}
\end{equation}
\begin{equation}
  \yhat(x)\defn s^0(x)-\widehat J(x)\widehat\beta.
  \label{EqnPredictor}
\end{equation}
\noindent
We select $\lambda$ by validation on calibration data,
keeping candidates from the same context in the
same split and choosing the largest penalty among exactly tied scores.

The coefficient on the baseline score in~\eqref{EqnPredictor} is
fixed at one. An input with zero measured response to every probe
therefore retains its baseline score. The coefficients are shared
across inputs, while each input's responses determine its correction.
Human annotations are needed to estimate $\widehat\beta$;
evaluation of a new input uses only the baseline and probed scores.
We also evaluate \alphaprism{}, with prediction
$\yhat(x)=\widehat\alpha s^0(x)-\widehat J(x)\widehat\beta$.
It fits $\alpha$ and $\beta$ jointly, penalizing only $\beta$
(\cref{AppCalibrationSensitivity}).

\myparagraph{Score encoding}
\emph{Single-candidate pointwise scoring} rates one candidate per call;
\emph{batched pointwise scoring} rates a context's candidates together
and can depend on their order, as \cref{FigAssumptions}(b) illustrates.
For within-context evaluation, we center scores and probe responses
and rescale training targets to the baseline score norm
(\cref{AppWithinContext}). Ranking-based encodings are described in
\cref{AppScoreAcquisition}.

\subsection{Assessing the probe assumptions}
\label{SecInterpretation}

We examine the probe measurements using SummEval, which
provides expert annotations for 16 summaries of each of 100 source
articles \citep{fabbri2021summeval}. The diagnostics below specify their
subsets and evaluation formats. Unless stated otherwise, they use
Jev-1.13, a model whose output is restricted to a probability
distribution over five ordered score levels; we take the expected level,
so scores are continuous rather than integers. Prediction results use
the mean within-document Spearman correlation with the average of the
four expert criteria, following the use of rank correlation in
LLM-judge evaluation \citep{liu2023geval,kola2026saja}. The order
diagnostic uses batched pointwise scoring of all 16 summaries.

\begin{figure}[!htbp]
  \centering
  \begin{minipage}[t]{0.40\linewidth}
    \centering
    \includegraphics[width=\linewidth]{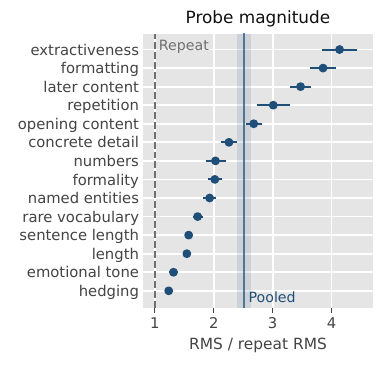}
    \par\smallskip {\small (a)}
  \end{minipage}\hfill
  \begin{minipage}[t]{0.28\linewidth}
    \centering
    \includegraphics[width=\linewidth]{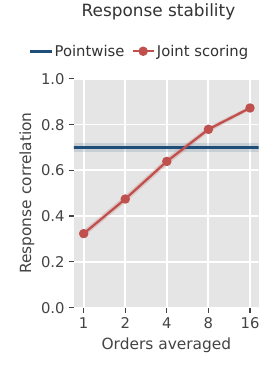}
    \par\smallskip {\small (b)}
  \end{minipage}\hfill
  \begin{minipage}[t]{0.28\linewidth}
    \centering
    \includegraphics[width=\linewidth]{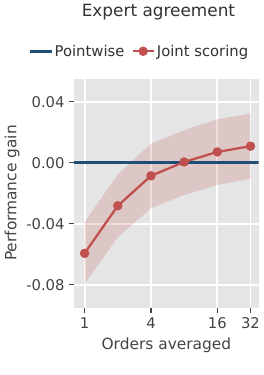}
    \par\smallskip {\small (c)}
  \end{minipage}
  \caption{Probe signal and presentation-order sensitivity for Jev on
  100 SummEval documents. (a) Probe RMS relative to Vanilla repeat
  variation; the vertical band pools properties. (b) Probe-response
  correlation between disjoint sets of averaged candidate orders, using
  length and formality probes of both signs. The blue reference uses
  repeated pointwise calls. (c) Uncalibrated Spearman gain over pointwise
  scoring. Joint scoring here is batched pointwise: one rating per
  candidate, in one call per order, versus 16 single-candidate calls
  per document; budgets differ. Bars and bands are 95\%
  document-bootstrap intervals. These diagnostics use separate runs
  from Table~\ref{TabExpertAgreement}; protocols appear in
  \cref{AppPresentationOrder,AppControlVariation}.}
  \label{FigAssumptions}
\end{figure}

\myparagraph{Signal above repeat variation}
\Cref{FigAssumptions}(a) compares probe displacements with differences
between repeated Vanilla scores, centering both within each document.
For Jev's 14-property bank, measured with
single-candidate scoring over three runs, RMS\footnote{For centered
displacements $d_1,\ldots,d_N$, the root mean square is
$\operatorname{RMS}(d)=\sqrt{N^{-1}\sum_{i=1}^N d_i^2}$.
We pool over candidates, documents, signs, and runs as appropriate.
The reference pools pairwise differences between Vanilla runs;
see \cref{AppControlVariation}.}
ratios range from $1.24$ for hedging to $4.14$ for extractiveness.
Every attribute's 95\% lower bound exceeds one: all 14 show movement
above observed Vanilla repeat variation. The pooled ratio is $2.51$
($[2.40,2.63]$), and mean across-run response
correlation is $0.76$ (\cref{FigRepeatReproducibility}). Response
magnitudes also differ across recorded judge settings, as the
quality-probe comparison with Sol shows (\cref{AppControlVariation}).

\myparagraph{Sensitivity to presentation order}
In batched pointwise scoring, candidate order affects both scores and
probe responses. In Jev's diagnostic, probe-response
correlation falls from $0.68$ between repeats of one order to $0.32$
across orders. \Cref{FigAssumptions}(b) shows that averaging eight
orders raises stability to $0.78$. \Cref{FigAssumptions}(c) shows the
corresponding rise in uncalibrated expert Spearman, from $0.587$ for
one order to $0.647$ for eight, matching the pointwise estimate.
Averaging common candidate orders
for baseline and probes therefore reduces presentation sensitivity;
it does not establish permutation invariance. Details appear in
\cref{AppPresentationOrder}.

\myparagraph{Response to weight instructions} We compare
\emph{slightly more/less}, \emph{more/less}, and \emph{much more/less}
weight instructions for six properties. For each attribute $k$, the
control $p_k^0$ requests its usual weight: ``exactly as heavily as you
normally would''. At each wording level, define the directional
half-difference and the shared displacement from this control:
\begin{align*}
  O_k(x)
  &\defn \frac{s(x;p_k^+)-s(x;p_k^-)}{2},
 \\
  E^{\mathrm{c}}_k(x)
  &\defn \frac{s(x;p_k^+)+s(x;p_k^-)}{2}-s(x;p_k^0).
\end{align*}
Here $O_k=[\widehat J^{\mathrm{half}}]_k$; symmetric score changes about
the control give $E^{\mathrm{c}}_k=0$.

\begin{figure}[!htbp]
  \centering
  \begin{minipage}[t]{0.485\linewidth}
    \centering
    \includegraphics[width=\linewidth]{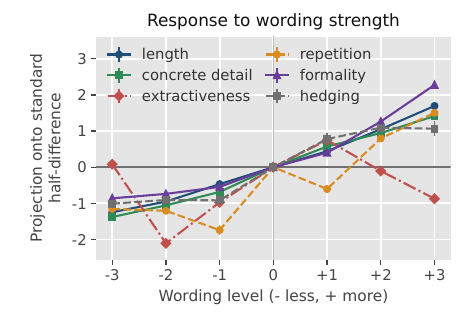}
    \par\smallskip {\small (a)}
  \end{minipage}\hfill
  \begin{minipage}[t]{0.485\linewidth}
    \centering
    \includegraphics[width=\linewidth]{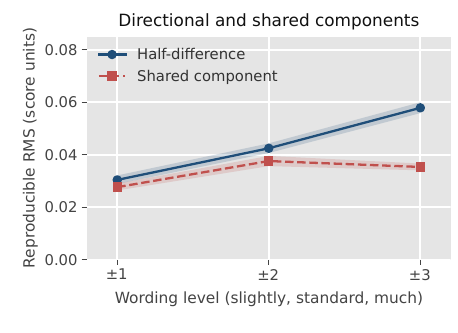}
    \par\smallskip {\small (b)}
  \end{minipage}
  \caption{Weight instructions for Jev on SummEval.
(a) Responses projected onto each attribute's standard half-difference.
Zero requests the usual weight; $\pm1$, $\pm2$, and $\pm3$ denote
\emph{slightly}, unmodified, and \emph{much} more/less weight.
These are wording categories, not numerical steps.
(b) Reproducible RMS of $O$ and $E^{\mathrm c}$, pooled over the same
six properties. Both panels use attribute-specific controls and
cross-products between ten repeated runs on the same 100 documents;
bars and bands are 95\% document-bootstrap intervals
(\cref{AppProbeEstimation}).}
  \label{FigProbeResponses}
\end{figure}

\Cref{FigProbeResponses}(a) shows projected responses across ten
replicate Jev runs: responses for length, concrete detail, and
formality increase with emphasis. Repetition changes projection sign
between \emph{slightly more} and \emph{more}, while hedging varies
little across the three less-weight instructions. Extractiveness
is non-monotone across wording levels.
\Cref{FigProbeResponses}(b) shows that, pooled over all six properties,
the directional component grows with wording strength. The shared
component remains substantial: its reproducible RMS is $0.91$,
$0.89$, and $0.61$ times the directional RMS at slight, standard,
and strong wording, respectively;
estimation details appear in
\cref{AppProbeEstimation}.

These patterns motivate using probe responses as empirical sensitivity
features. Under a smooth local weight model, $O$ approximates the score
Jacobian applied to the induced perturbation when the two prompts
produce small, symmetric weight changes, connecting the probes to
local sensitivity analysis and surrogate explanations
\citep{cook1986local,ribeiro2016trust}. Wording levels specify no
numerical step size, so we use the measured contrasts directly.
\Cref{AppResponseBaselines} gives the local model and relates the
controls to the Vanilla calibration baseline; \cref{SecAblations}
tests the predictive value of retaining both responses.

\section{Calibration performance and interpretation}
\label{SecResults}

\myparagraph{Evaluation protocol}
SummEval provides 100 documents with 16 summaries each. Our target
averages expert coherence, consistency, fluency, and relevance ratings.
HANNA provides 96 prompts with 11 stories each, rated by three people
on six criteria. Our main metric is Spearman correlation, also used
for LLM-judge evaluation by Liu et al.~\citep{liu2023geval}
and Kola et al.~\citep{kola2026saja}.
We compute correlations within documents (SummEval) or prompts (HANNA)
and average across contexts to assess agreement with human rankings.
We evaluate fitted methods on the same five held-out groups of
documents or prompts,
training on the remaining groups. We choose \prism{}'s regularization
by validation within the training data using the same metric;
exactly tied scores favor stronger regularization. Individual and
paired 95\% intervals use 2,000 document/prompt-bootstrap samples.
In the main SummEval comparison,
\prism{} uses single-candidate expected ratings for Jev and batched
pointwise ratings for Sol; both use single-candidate scoring on HANNA.
Full acquisition and calibration protocols, including the additional
Sonnet and Gemma settings, appear in
\cref{AppScoreAcquisition,AppWithinContext}.
Model versions and inference settings are collected in
\cref{AppModelSettings}.

\subsection{Agreement with reference judgments}
\label{SecExpertAgreement}

\Cref{TabExpertAgreement} compares Jev and Sol on summary evaluation
(SummEval) and story evaluation (HANNA). We compare Vanilla and the
two \prism{} variants with rubric-based methods: SAJA combines broad
or granular criterion ratings using a random forest
\citep{kola2026saja}, while we average normalized G-Eval criterion scores
\citep{liu2023geval}. Full results for all four judges and baseline
protocols appear in \cref{TabSummEvalAgreement,TabHannaAgreement,AppGEval}.

\begin{table}[!htbp]
  \centering
  \small
  \setlength{\tabcolsep}{10pt}
  \renewcommand{\arraystretch}{1.1}
  \begin{tabular}{@{}lcc@{\hspace{2.5em}}cc@{}}
    \toprule
    & \multicolumn{2}{c}{SummEval} & \multicolumn{2}{c}{HANNA}\\
    \cmidrule(lr){2-3}\cmidrule(l){4-5}
    Method & Jev & Sol & Jev & Sol\\
    \midrule
    \multicolumn{5}{@{}l}{\emph{Rubric-based scoring}}\\[2pt]
    SAJA broad & \shortstack{$0.703$\\[-1pt]{\footnotesize$[0.672,0.732]$}} & \shortstack{$0.659$\\[-1pt]{\footnotesize$[0.627,0.688]$}} & \shortstack{$0.524$\\[-1pt]{\footnotesize$[0.484,0.565]$}} & \shortstack{$0.443$\\[-1pt]{\footnotesize$[0.390,0.492]$}}\\[3pt]
    SAJA granular & \shortstack{$\mathbf{0.745}$\\[-1pt]{\footnotesize$[0.719,0.770]$}} & \shortstack{$0.707$\\[-1pt]{\footnotesize$[0.679,0.732]$}} & \shortstack{$0.549$\\[-1pt]{\footnotesize$[0.508,0.590]$}} & \shortstack{$0.528$\\[-1pt]{\footnotesize$[0.482,0.574]$}}\\[3pt]
    G-Eval & \shortstack{$0.701$\\[-1pt]{\footnotesize$[0.670,0.729]$}} & \shortstack{$\mathbf{0.721}$\\[-1pt]{\footnotesize$[0.696,0.745]$}} & \shortstack{$0.528$\\[-1pt]{\footnotesize$[0.483,0.575]$}} & \shortstack{$\mathbf{0.587}$\\[-1pt]{\footnotesize$[0.543,0.632]$}}\\[3pt]
    \midrule
    \multicolumn{5}{@{}l}{\emph{Rubric-free scoring}}\\[2pt]
    Vanilla & \shortstack{$0.654$\\[-1pt]{\footnotesize$[0.615,0.691]$}} & \shortstack{$0.693$\\[-1pt]{\footnotesize$[0.661,0.722]$}} & \shortstack{$0.459$\\[-1pt]{\footnotesize$[0.409,0.505]$}} & \shortstack{$0.462$\\[-1pt]{\footnotesize$[0.428,0.497]$}}\\[3pt]
    \rowcolor{gray!10}
    \prism{} (ours) & \shortstack{$0.705$\\[-1pt]{\footnotesize$[0.673,0.735]$}} & \shortstack{$0.706$\\[-1pt]{\footnotesize$[0.674,0.734]$}} & \shortstack{$\underline{0.599}$\\[-1pt]{\footnotesize$[0.558,0.639]$}} & \shortstack{$0.546$\\[-1pt]{\footnotesize$[0.502,0.588]$}}\\[3pt]
    \rowcolor{gray!10}
    \alphaprism{} (ours) & \shortstack{$\underline{0.712}$\\[-1pt]{\footnotesize$[0.683,0.741]$}} & \shortstack{$\underline{0.712}$\\[-1pt]{\footnotesize$[0.682,0.740]$}} & \shortstack{$\mathbf{0.603}$\\[-1pt]{\footnotesize$[0.562,0.642]$}} & \shortstack{$\underline{0.557}$\\[-1pt]{\footnotesize$[0.514,0.598]$}}\\[3pt]
    \bottomrule
  \end{tabular}
  \caption{Agreement with human judgments on SummEval and HANNA.
  Cells show mean within-document (SummEval) or within-prompt (HANNA)
  Spearman correlation (higher is better), with 95\% document/prompt-bootstrap
  intervals. Bold and underline mark the best and second-best displayed
  means, including ties. Groups distinguish scoring protocols, not probe
  provenance; \prism{} fixes $\alpha=1$ and \alphaprism{} learns it.
  Full four-judge results appear in \cref{TabSummEvalAgreement,TabHannaAgreement}.}
  \label{TabExpertAgreement}
\end{table}

The strongest method depends on the judge and task. Across all four
settings in \cref{TabExpertAgreement}, \prism{} and \alphaprism{}
significantly improve on Vanilla, based on paired 95\% bootstrap
intervals, while achieving competitive agreement with the rubric-based
methods. They do so using only overall scores and probe responses,
without criterion-level scoring. \Cref{TabProbeAttributes} further
supports this distinction: replacing rubric-derived quality probes
with style probes yields similar performance in the matched
Jev--SummEval comparison (\cref{SecAblations}).

\myparagraph{Additional benchmarks}
We further evaluate Jev on dialogue quality (Topical-Chat)
\citep{mehri2020usr}, helpfulness (HelpSteer2)
\citep{wang2024helpsteer2}, conversational preferences (MT-Bench)
\citep{zheng2023judging}, and correctness (JudgeBench)
\citep{tan2025judgebench}. For each task, probe screening,
calibration, and regularization selection use a development set;
evaluation uses separate held-out contexts. We report
\emph{pairwise accuracy} (hereafter accuracy): the fraction of pairs
ordered as in the reference judgments, averaged within and then
across contexts. Reference ties are excluded; predicted ties receive
half credit. Full protocols appear in \cref{AppCrossBenchmark}.

\begin{table}[!htbp]
  \centering
  \small
  \setlength{\tabcolsep}{8pt}
  \renewcommand{\arraystretch}{1.10}
  \begin{tabular}{@{}lrcccc@{}}
    \toprule
    & & \multicolumn{2}{c}{Rubric-based} & \multicolumn{2}{c}{Rubric-free}\\
    \cmidrule(lr){3-4}\cmidrule(l){5-6}
    Benchmark & $n$ & SAJA broad & SAJA granular & Vanilla & \prism{}\\
    \midrule
    Topical-Chat & 34 & $0.799$ & $0.831$ & $\underline{0.835}$ & $\mathbf{0.837}$\\
    HelpSteer2 & 188 & $0.665$ & $0.710$ & $\underline{0.739}$ & $\mathbf{0.755}$\\
    MT-Bench & 81 & $0.808$ & $0.805$ & $\underline{0.818}$ & $\mathbf{0.828}$\\
    JudgeBench & 287 & $0.676$ & $0.707$ & $\underline{0.789}$ & $\mathbf{0.791}$\\
    \bottomrule
  \end{tabular}
  \caption{Accuracy $\uparrow$ on additional benchmarks, using Jev.
  Groups follow \cref{TabExpertAgreement}.
  \prism{} uses every signal-screened unit from the expanded pool.
  $n$ counts confirmation contexts with non-tied reference pairs;
  predicted ties receive half credit. Bold and underline mark the best
  and second-best displayed values, including rounding ties. All variants and paired comparisons
  appear in \cref{AppCrossBenchmark}.}
  \label{TabBeyondSummEval}
\end{table}

\Cref{TabBeyondSummEval} shows that Vanilla already exceeds both SAJA
variants on all four tasks, including the granular rubric. With
regularization chosen on development data, \prism{} yields small
further improvements in all four point estimates. The paired 95\%
gain intervals include zero (\cref{TabXBenchContrasts}).

\subsection{Correction profiles across judges}
\label{SecCorrectionProfiles}

\begin{figure}[!htbp]
  \centering
  \begin{minipage}[t]{0.32\linewidth}
    \centering
    \includegraphics[width=\linewidth]{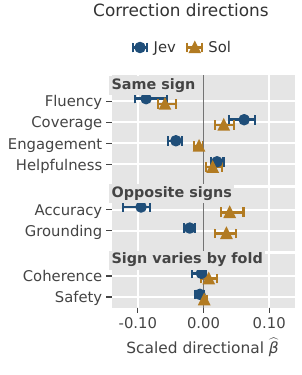}
    \par\smallskip {\small (a)}
  \end{minipage}\hfill
  \begin{minipage}[t]{0.32\linewidth}
    \centering
    \includegraphics[width=\linewidth]{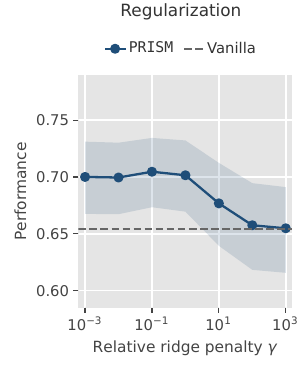}
    \par\smallskip {\small (b)}
  \end{minipage}\hfill
  \begin{minipage}[t]{0.32\linewidth}
    \centering
    \includegraphics[width=\linewidth]{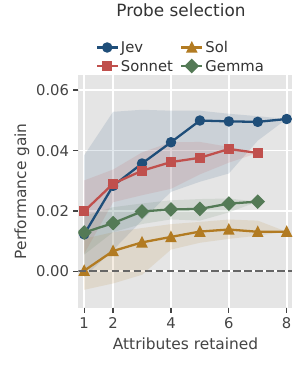}
    \par\smallskip {\small (c)}
  \end{minipage}
  \caption{Correction profiles and robustness on SummEval.
  (a) Normalized directional components of $\widehat\beta$ for Jev/Sol;
  bars span the five
  overlapping training folds, not confidence intervals. Positive
  coefficients subtract the response. (b) Jev's agreement at fixed
  relative penalties $\gamma$ ($\lambda$ multiplied by a training-fold
  constant); bands are
  95\% document-bootstrap intervals. Main results select penalties
  inside training folds. (c) Median gains over Vanilla across all
  probe subsets of each size; bands are 10th--90th percentiles across
  subsets, not confidence intervals. Dashed lines mark Vanilla.}
  \label{FigCorrectionProfiles}
\end{figure}

\myparagraph{Interpreting the learned coefficients}
The fitted coefficients $\widehat\beta$ give an interpretable
approximation to scoring bias in the chosen probe-response coordinates:
subtracting $\widehat J\widehat\beta$ corrects the score toward human
judgments. \Cref{FigCorrectionProfiles}(a) plots their directional
components, $\widehat\beta_k^++\widehat\beta_k^-$, scaled by the probe
response magnitude relative to Vanilla (\cref{AppCorrectionProfiles}).
Positive values subtract the corresponding response; negative values add it.

\myparagraph{Shared and opposing corrections}
\Cref{FigCorrectionProfiles}(a) shows that the inferred bias differs
in both magnitude and direction across judges. Jev and Sol both add
the fluency response, but their corrections for accuracy and grounding
point in opposite directions. Profiles for Sonnet and Gemma
appear in \cref{AppCorrectionProfiles}.

\subsection{Ablations of the correction}
\label{SecAblations}

\myparagraph{Retaining both probe responses}
\Cref{TabCalibrationAblations} compares the half-difference, paired
responses, and a diagonal Taylor correction. Paired \prism{} retains
the shared score changes discarded by the half-difference, including
second-difference information. Its point estimate improves for Jev
and is nearly unchanged for Sol; both paired 95\% gain intervals
include zero. The explicit Taylor constraint adds little in either
case (\cref{AppResponseAblation}).

\myparagraph{Robustness to probe attributes}
Calibration remains effective when probes omit explicit rubric
criteria. For Jev on SummEval, we compare two non-overlapping banks
of eight attributes: rubric-derived quality criteria, such as
accuracy and coherence, and style properties, such as length and
extractiveness. \Cref{TabProbeAttributes} shows similar agreement for
the two banks and little benefit from combining them. This supports rubric-free
probe construction: useful correction signals need not come from
reproducing the rubric's criteria in the probe bank. The matched
protocol and full attribute lists appear in \cref{AppProbeSelection}.

\begin{table}[!htbp]
  \centering
  \begin{subtable}[t]{0.51\linewidth}
    \centering\small
    \setlength{\tabcolsep}{4pt}
  \begin{tabular}{@{}lcc@{}}
    \toprule
    Representation & Jev & Sol\\
    \midrule
    Half-difference & \shortstack{$0.696$\\[-1pt]{\scriptsize$[0.664,0.727]$}} & \shortstack{$0.707$\\[-1pt]{\scriptsize$[0.677,0.734]$}}\\[3pt]
    Paired \prism{} & \shortstack{$0.705$\\[-1pt]{\scriptsize$[0.673,0.735]$}} & \shortstack{$0.706$\\[-1pt]{\scriptsize$[0.674,0.734]$}}\\[3pt]
    Diagonal Taylor & \shortstack{$0.705$\\[-1pt]{\scriptsize$[0.674,0.736]$}} & \shortstack{$0.707$\\[-1pt]{\scriptsize$[0.677,0.735]$}}\\[3pt]
    \bottomrule
  \end{tabular}
    \caption{Response representations}
    \label{TabCalibrationAblations}
  \end{subtable}\hfill
  \begin{subtable}[t]{0.46\linewidth}
    \centering\small
    \setlength{\tabcolsep}{5pt}
    \renewcommand{\arraystretch}{1.45}
  \begin{tabular}{@{}lcc@{}}
    \toprule
    Probe attributes & \prism{} & 95\% CI\\
    \midrule
    Quality (8) & $0.705$ & $[0.673,0.735]$\\
    Style (8) & $0.710$ & $[0.679,0.738]$\\
    Combined (16) & $0.711$ & $[0.681,0.740]$\\
    \bottomrule
  \end{tabular}
    \caption{Probe attributes}
    \label{TabProbeAttributes}
  \end{subtable}
  \caption{Ablations on SummEval: mean within-document Spearman and
  95\% document-bootstrap intervals.
  (a) All representations use the same calls and folds, with the
  Vanilla coefficient fixed at one. Diagonal Taylor fits a second-order
  correction; paired \prism{} already contains its second-difference
  inputs (\cref{AppResponseAblation}).
  (b) Jev with pointwise scoring. Parentheses give attribute counts;
  Combined is the union. Predictor Vanilla scores, folds, and calibration
  are held fixed.}
  \label{TabCorrectionAblations}
\end{table}

\myparagraph{Probe selection}
\Cref{FigCorrectionProfiles}(c) shows typical gains rising and then
leveling off as attributes are added. Each point is the median over
\emph{every} subset of that size, with regularization selected
separately on training data, rather than a few hand-picked subsets.
The leveling suggests overlapping coverage of useful correction
directions: once several attributes are included, additional ones
contribute less. Mean gains also survive every single-attribute
omission (\cref{AppProbeSelection}), consistent with the probe-bank
robustness in \cref{TabProbeAttributes}.

\myparagraph{Ridge regularization sensitivity}
\Cref{FigCorrectionProfiles}(b) shows that Jev's agreement is stable
across a range of moderate penalties. We plot the relative penalty
$\gamma$($\lambda$ multiplied by a training-fold
  constant)\footnote{For a training fold with $n_T$ examples and response
matrix $\mathbf J_T\in\real^{n_T\times q}$, we use
$\gamma=\lambda n_Tq/\|\mathbf J_T\|_F^2$.
Thus, $\gamma$ is $\lambda$ multiplied by a constant for that training
design; only the parameterization changes.}
on the horizontal axis. When $\gamma\gg1$, the correction shrinks
toward zero and agreement approaches the Vanilla level.
Reported predictors select regularization on training data; full
sweeps and the validation tie rule appear in
\cref{AppCalibrationSensitivity,TabXBenchTiePolicy}.

\section{Conclusion}
\prism{} connects perturbation analysis to rubric-free LLM evaluation.
It represents changes in overall scores under attribute-emphasis
instructions as a surrogate Jacobian and learns a ridge correction
from human labels. Inference uses only baseline and probed scores,
without requesting criterion-level judge ratings. On SummEval and
HANNA, \prism{} and \alphaprism{} significantly improve on Vanilla
in the main Jev and Sol comparisons while achieving ranking agreement
competitive with rubric-based SAJA and G-Eval. The matched
Jev--SummEval ablation yields similar agreement from rubric-derived
quality probes and style probes, supporting calibration without
reproducing rubric criteria. Learned coefficients further expose
shared and opposing correction directions across judges, making
the use of probe responses in the correction interpretable.
\label{MainContentEnd}

\section*{Limitations}
We study rubric-free calibration of text-based judgments, using human
labels to fit a correction for each judge. Additional probe calls
incur inference cost, motivating comparisons under matched budgets
and adaptive probe selection. Transfer across tasks, model versions,
and rater populations remains to be tested. Benchmark contamination
cannot be ruled out; evaluation on newly collected data would
strengthen the evidence for generalization.

Our experiments calibrate overall scores for fixed inputs. A next
step is to target biases arising from joint candidate presentation
in batched pointwise scoring or listwise ranking, using instruction
probes that preserve candidate content.

\section*{Acknowledgments}
KY was
supported by the Takenaka Scholarship Foundation.
We used ChatGPT/Codex and Claude Code for writing and coding
assistance. The authors take responsibility for the content and
results.

\clearpage
\appendix
\begin{center}
  \large\bfseries
  Appendix for
  ``PRISM: Calibrating Black-Box Judges with Counterfactual Probes''
\end{center}
\vspace{1em}
\section{Experimental setup and implementation details}
\label{AppCalibrationProtocol}
This section collects the acquisition, calibration, and evaluation
settings used in \cref{SecInterpretation,SecResults}.

\subsection{Model versions and inference settings}
\label{AppModelSettings}

\Cref{TabModelSettings} consolidates the recorded model identifiers
and inference settings for the main comparisons and supplementary
experiments. Jev requests use the alias \texttt{typesafe/jev-1.13};
the returned version is \texttt{jev-1.13-20260917}. The other logs
provide the identifiers shown, without a finer dated snapshot.
An omitted effort field is distinguished from explicitly disabling
reasoning.

\begin{table}[!htbp]
  \centering
  \small
  \setlength{\tabcolsep}{5pt}
  \renewcommand{\arraystretch}{1.15}
  \begin{tabular}{@{}lp{0.35\textwidth}lcc@{}}
    \toprule
    Judge & Recorded model identifier & Reasoning effort &
    \shortstack{Temperature\\Vanilla/\prism{}/SAJA} &
    \shortstack{Temperature\\G-Eval}\\
    \midrule
    Jev & \texttt{typesafe/jev-1.13-20260917} & Not set & -- & --\\
    Sonnet & \texttt{anthropic/claude-sonnet-4.5} & Not set & $0$ & $1$\\
    Sol & \texttt{gpt-5.6-sol} & \texttt{none} & $0$ & $1$\\
    Gemma & \texttt{google/gemma-4-31b-it} & Not set & $0$ & $1$\\
    \bottomrule
  \end{tabular}
  \caption{Model versions and requested inference settings across
  experiments. ``Not set'' and -- denote omitted API fields, not
  confirmed provider defaults. Sol's OpenRouter identifier adds the
  prefix \texttt{openai/}. Jev returns typed-rating probabilities;
  the other judges use chat completions with \texttt{top\_p=1}.
  Probe diagnostics use the corresponding Vanilla/\prism{} settings.}
  \label{TabModelSettings}
\end{table}

\myparagraph{Interfaces and serving providers}
Jev uses the TypeSafe Decisions API. Sonnet uses Anthropic through
OpenRouter. Sol uses OpenRouter for SummEval \prism{} and SAJA,
and the direct OpenAI API for HANNA and G-Eval; reasoning effort is
explicitly \texttt{none} in both interfaces. Gemma uses OpenRouter
with \texttt{bf16} requested: Venice serves SummEval \prism{}, SAJA,
and their wording diagnostics; Novita serves HANNA \prism{} and
SAJA and SummEval G-Eval. HANNA G-Eval responses come from Novita
and Crusoe. These serving differences are part of the recorded
experimental settings.

\myparagraph{Output budgets}
SummEval Sonnet/Gemma probe calls allow 900 output tokens. Sol's
single-candidate and batched pointwise calls allow 40 and 400,
respectively, with 120 for the four-criterion diagnostic.
SummEval chat-based SAJA allows 700 tokens for broad and 1,400
for granular ratings. HANNA chat-based probe and SAJA calls allow
400. G-Eval generally allows 20 tokens per completion; HANNA
Sonnet also includes calls capped at eight. Jev requests set no
output-token budget. Repetition counts and score formats are given
with each protocol below.

\myparagraph{Recorded computation}
Total recorded API usage for SummEval and the additional benchmarks
is approximately 1.4 million responses, 1.9 billion input tokens,
and 260 million output tokens, including development and diagnostic
experiments. These are aggregate research costs, not matched
per-method budgets, and exclude unlogged requests, local analysis
time, and AI-assistant use. Inference is API-hosted; provider GPU hours and parameter counts
are not recorded. Calibration and analysis use CPU-based numerical
routines on a local Mac without updating judge weights; cumulative
local runtime is unavailable.

\myparagraph{Software environment}
The analysis environment inspected on 13 October 2026 uses Python
3.14.6 on macOS/ARM64, with a 10-core Apple M5 CPU and 24\,GB of
memory. The principal libraries are
\href{https://numpy.org}{NumPy}~2.5.3,
\href{https://scipy.org}{SciPy}~1.18.1,
\href{https://scikit-learn.org}{scikit-learn}~1.9.1,
\href{https://pandas.pydata.org}{pandas}~3.0.5, and
\href{https://matplotlib.org}{Matplotlib}~3.11.2.
Data loading and requests use \texttt{datasets}~5.0.1,
\texttt{huggingface-hub}~1.30.0, \texttt{pyarrow}~25.0.1,
and \texttt{httpx}~0.28.1.
These versions describe the analysis environment; acquisition
environments were not all version-locked.

\subsection{Scoring protocols and probe banks}
\label{AppScoreAcquisition}
\label{AppHannaComparison}

The four primary settings use the same 1,600 document--summary pairs
and the overall expert target. Jev-1.13 returns an expected score
over five categories, coded $0$--$4$, through the Decisions API.
Its individual-summary call contains all 38 scoring questions: one
baseline, two wordings for each sign of eight attributes, and five
neutral clauses. Sol returns integer ratings on a $1$--$10$ scale,
jointly scoring the 16 summaries for each instruction. Sonnet and
Gemma return rankings with ties; scores are negative midranks,
aligned by summary identity and averaged across successful
repetitions. Sonnet uses the overall-quality run, rather than an
average of criterion-specific coherence and relevance runs.

\begin{table}[!htbp]
  \centering
  \small
  \setlength{\tabcolsep}{8pt}
  \begin{tabular}{@{}llrrr@{}}
    \toprule
    Judge & Probe-score format & Attributes & Conditions & Calls/document\\
    \midrule
    Jev & Individual expected ratings & $8$ & $38$ & $16$\\
    Sonnet & Joint rankings; two orders & $7$ & $34$ & $68$\\
    Sol & Joint integer ratings & $8$ & $38$ & $38$\\
    Gemma & Joint rankings; two orders, two blocks & $7$ & $34$ & $136$\\
    \bottomrule
  \end{tabular}
  \caption{Nominal acquisition designs for the score conditions used
  in the calibration analysis. Conditions include five neutral
  clauses, which are not columns of the paired-response predictor.
  Additional surface perturbations and Section~2 diagnostics are
  excluded. Each SAJA variant uses one rubric call per summary,
  or 16 calls per document. Query and token budgets are not matched.}
  \label{TabCalibrationDesigns}
\end{table}

The quality bank is adapted from SAJA's broad rubric, replacing its
overall-score item with coverage; it comprises accuracy, grounding,
coverage, coherence, fluency, engagement, helpfulness, and safety.
These attributes change the overall evaluation rather than elicit
separate criterion ratings. Exact attribute clauses for the matched
quality/style comparison are listed in \cref{TabProbeBankDefinitions}. The seven-attribute
style bank comprises length, detail, repetition, specificity, lead,
polish, and confidence. The separate Jev style-bank comparison uses
length, specificity, lead, polish, repetition, confidence,
extractiveness, and structure. These are distinct recorded banks;
matching an attribute name alone does not match an elicitation
protocol.

\myparagraph{Recorded bank coverage}
The primary settings do not exhaust the available runs.
\Cref{TabProbeBankCoverage} distinguishes those used in the main
comparison from additional recorded banks. Jev's eight-attribute
style bank shares six attributes with the seven-attribute bank:
it lacks detail and instead includes extractiveness and structure.
A shared subset of names permits an exploratory comparison,
but does not remove differences in wording or score elicitation.

\begin{table}[!htbp]
  \centering
  \small
  \setlength{\tabcolsep}{9pt}
  \begin{tabular}{@{}lll@{}}
    \toprule
    Judge & Quality bank (8 attributes) & Style bank (7 attributes)\\
    \midrule
    Jev & Primary: expected ratings & Different bank: 8 attributes, 6 shared\\
    Sonnet & Additional: joint ratings & Primary: joint rankings\\
    Sol & Primary: joint ratings & Additional: joint rankings\\
    Gemma & Not found in inspected records & Primary: joint rankings\\
    \bottomrule
  \end{tabular}
  \caption{Probe-bank coverage in the inspected SummEval records.
  Primary denotes the source of the main result and correction profile;
  additional denotes an existing run outside that comparison. All listed
  runs cover 100 documents. These records do not constitute a matched
  bank-by-judge experiment; presentation and repetition counts also differ.}
  \label{TabProbeBankCoverage}
\end{table}

\myparagraph{HANNA}
The full-data comparison uses 96 prompts with 11 stories each. Human
targets average three raters across six criteria. A common bank has
16 paired property probes and eight reading-context probes, giving
40 response columns without screening. Jev returns expected ratings
for all conditions in one story-level call; Sol and Gemma return a
1--10 integer in each story--condition call. Sonnet uses 0--100 ratings with corresponding anchors.
A matched 1--10 comparison holds the model, bank, and folds fixed
(\cref{AppScoreResolution}). Sol's supplementary
probability-weighted probe scores do not replace its integer run.
The separate 53-prompt HANNA confirmation study is described in
\cref{AppCrossBenchmark}.

\myparagraph{Missing probe scores}

The recorded matrices have no missing cells for Jev, and 64, 16,
and 144 missing cells for Sonnet, Sol, and Gemma, respectively,
after averaging available responses. Each of these missing blocks
comprises all 16 candidates of a document under one instruction.
Missing values are replaced by a column mean.
For the centered probe predictors, any common value within such
a block cancels, so these predictors do not depend on the value
of that mean. This does not reconstruct the missing responses.

For HANNA, the evaluator uses score-column means to impute failed
replies before fitting: 694 of 48,576 scores for Sonnet's
1--10 condition, 1,076 of 48,576 for its 0--100 condition, and 179 of 48,576
for Gemma's 1--10 run. These are recorded missing-data treatments,
not additional judge measurements.

\subsection{Calibration and uncertainty}
\label{AppWithinContext}
For experiments that predict candidate order within a context, we
center the baseline scores and response columns within each group.
For a group $g$ of $n_g$ candidates, let
$C_g\defn I_{n_g}-\boldsymbol 1\boldsymbol 1^\top/n_g$.
The regression uses $C_g\mathbf s_g^0$ and
$C_g\widehat{\mathbf J}_g$, with the human target rescaled to match
the baseline's within-group norm:
\begin{align}
  \mathbf y_g^*
  \defn
  \frac{\|C_g\mathbf s_g^0\|_2}
       {\|C_g\mathbf y_g\|_2}\,C_g\mathbf y_g.
  \label{EqnRankingTarget}
\end{align}
The target is zero when the denominator is zero. At deployment, a
group's baseline and responses can be centered without human labels.
The resulting scores recover an ordering and need not lie on the
original human rating scale. For listwise ranking encodings, we align
outputs by candidate identity and average scores over the prescribed
candidate orders before forming the probe responses.

For SummEval, each outer fold holds out 20 documents and fits on the other 80.
The folds are obtained by shuffling sorted document identifiers
with seed 42 and splitting into five equal parts. Inner validation
uses five grouped folds on the 80 training documents and selects
the penalty by mean within-document Spearman correlation. Both
baseline and response columns are centered within documents;
the training expert target is transformed by
\eqref{EqnRankingTarget}. Expert labels from the outer test fold
are used only for evaluating its predictions.

HANNA uses the same five-fold outer/inner procedure, grouping by its
96 prompts (seed 42); a constant prediction within a prompt receives
Spearman correlation zero. Exact validation-score ties favor the
larger penalty in both benchmarks.

The bootstrap uses 2,000 paired resamples of the out-of-fold contexts:
100 documents for SummEval and 96 prompts for HANNA. It takes
percentile intervals without refitting the regressions. Correlations
are averaged across resampled contexts. Differences use the same resampled
contexts for both methods. Intervals are unadjusted for multiple comparisons
and are conditional on the recorded outputs and fold assignment.
Although each fitted predictor is evaluated on excluded documents,
SummEval was also used during method and probe development.
These comparisons consequently assess the recorded designs;
an independent benchmark is needed to assess generalization
after that development process.

\subsection{Acquisition of comparison scores}
\label{AppBaselineAcquisition}
\label{AppGEval}
All SAJA rows use a 300-tree random forest (seed 42), fitted on the
same training contexts as \prism{}. SummEval broad/granular inputs
use expected 0--4 ratings for Jev, 1--10 integers for Sol, and
1--5 integers for Sonnet/Gemma. HANNA uses 1--10 ratings for all
judges, with Jev supplying expectations over ten categories.
Refits use one worker for reproducible tree summation. HANNA
mean-imputes 280/8,448 broad and 476/29,568 granular scores for
Sonnet, and 8/8,448 broad scores for Gemma; its granular run and
both Jev/Sol runs require no SAJA imputation.

G-Eval \citep{liu2023geval} scores candidates per criterion.
We normalize and average these scores for our overall target,
without a fitted prediction head. SummEval uses the official
\emph{detailed} prompts at revision
\texttt{6f84404}: fluency is mapped from $[1,3]$ and coherence,
consistency, and relevance from $[1,5]$ to $[0,1]$. HANNA uses its
six criteria---relevance, coherence, empathy, surprise, engagement,
and complexity---with the original definitions and 1--5 anchors
\citep{chhun2022hanna}. Its evaluation steps were generated once
by Sol at temperature zero with effort \texttt{none}, then frozen
across judges before scoring. Human ratings are not included in
either benchmark's prompts.

\Cref{TabComparisonAcquisition} summarizes API-specific scoring.
Chat completions use temperature one and $\mathrm{top\_p}=1$;
valid ratings are averaged in draw order. Sol's SummEval calls use
a user message and a 20-token limit; the parser accepts a leading
rating or a rating following the criterion label. Probability-based
expectations renormalize over valid rating tokens. For HANNA,
Sonnet uses a score-label assistant prefill; Sol and Gemma receive
a system instruction identifying the supplied story as the item
to score and requesting only a rating. The criteria and anchors
remain unchanged.

\begin{table}[!htbp]
  \centering\small
  \setlength{\tabcolsep}{5pt}
  \renewcommand{\arraystretch}{1.12}
  \begin{tabular}{@{}lp{0.43\textwidth}p{0.40\textwidth}@{}}
    \toprule
    Judge & SummEval: 6,400 criterion scores & HANNA: 6,336 criterion scores\\
    \midrule
    Jev & Expected rating from typed probabilities; no imputation.
        & Expected rating from typed probabilities; no imputation.\\[3pt]
    Sonnet & First 10 valid draws, at most 20 attempts per item.
        All scores available; 97 items average 5--9 valid draws.
        & First 10 valid draws per item; all items complete.\\[3pt]
    Sol & Top-5 token probabilities for 6,229 items; sampling fallback
        for 168 (up to 20 valid draws in 40 attempts); 3 midpoint imputations.
        Of sampled items, 53 reach the cap below 20 valid draws.
        & Score-token probabilities for all items; no sampling fallback
        or imputation.\\[3pt]
    Gemma & Top-20 token probabilities, with up to 5 retries;
        fallback to 10 valid draws in 20 attempts.
        2,128 midpoint imputations, including 1,525 fluency items.
        & First 10 valid draws per item; all items complete.\\
    \bottomrule
  \end{tabular}
  \caption{Recorded G-Eval acquisition protocols. Midpoint imputation
  fills items with no valid score; it does not recover the missing
  judgment. These are model-specific adaptations of the published
  procedure, with unmatched sampling budgets across methods.}
  \label{TabComparisonAcquisition}
\end{table}

Gemma's SummEval failures frequently request a summary instead of
rating the supplied one; 33.3\% of its criterion scores are imputed.
For HANNA, all reported Sol and Gemma G-Eval scores use the
response-format instructions described above, with no missing items.

\subsection{Additional benchmark protocols}
\label{AppCrossBenchmark}
This study covers the four additional benchmarks in
\cref{TabBeyondSummEval} and a separate HANNA confirmation experiment.
The latter uses a development/confirmation split and an expanded
probe pool, distinct from the full-dataset comparison in
\cref{TabExpertAgreement}.

\myparagraph{Targets and units}
All five benchmarks use Jev-1.13 through the Decisions API, with
expected scores over five ordered levels on the $0$--$4$ scale.
HANNA contains 96 prompts with 11 stories each, including the human
story; the target averages three raters over six criteria. USR's
Topical-Chat annotations contain 60 contexts with six responses
each, using the mean Overall rating. HelpSteer2 uses the validation
split, with 519 requests and two responses per request, targeting
helpfulness. For these graded datasets, every within-context pair
with unequal target scores contributes an evaluation label.
MT-Bench contains 160 question--turn contexts and responses from
six models. Its training target is a response's human win rate;
evaluation uses individual non-tie human comparisons. For JudgeBench,
we combine 350 GPT-4o and 270 Claude-3.5-Sonnet pairs (620 total),
with binary correctness targets. Thus, agreement with human ratings and agreement with
verified correctness are distinct targets in this comparison.

\myparagraph{Splits and uncertainty}
A fixed hash of the benchmark and context identifier assigns each
context to exploration or confirmation. Probe screening, predictor
fitting, and hyperparameter selection use exploration data.
Reference ties are omitted from evaluation, predicted ties receive
$1/2$ credit, and accuracy is averaged within each context
before averaging eligible contexts. All exploration candidates,
including those with tied labels, remain in the regression fit.
The intervals resample eligible confirmation contexts with
replacement 2,000 times, using the same draws for paired contrasts.
They condition on the fitted predictors and do not include
uncertainty from selecting the probe bank or development protocol.

\begin{table}[!htbp]
  \centering
  \small
  \setlength{\tabcolsep}{7pt}
  \renewcommand{\arraystretch}{1.10}
  \begin{tabular}{@{}lrrrrr@{}}
    \toprule
    Benchmark & Explore & Confirm & Eval. pairs & Units & Columns\\
    \midrule
    HANNA & 43 & 53 & 2860 & 74 & 126\\
    Topical-Chat & 26 & 34 & 468 & 82 & 134\\
    HelpSteer2 & 251 (185) & 268 (188) & 188 & 85 & 137\\
    MT-Bench & 79 & 81 & 1280 & 79 & 131\\
    JudgeBench & 333 & 287 & 287 & 84 & 136\\
    \bottomrule
  \end{tabular}
  \caption{Recorded splits and primary response dimensions. Parentheses give the eligible contexts after excluding reference ties from pairwise evaluation; all other contexts contribute evaluation pairs. MT-Bench counts individual human judgments, so a response pair may contribute multiple labels. All candidates have parsed scores in every primary measurement bank.}
  \label{TabXBenchDesign}
\end{table}

The 24-unit and expanded 89-unit probe pools use the same
exploration/confirmation split. These benchmarks also inform method
development, so the comparisons assess the resulting designs rather
than validation on a new sample. MT-Bench splits by question--turn rather than by
question: 43 questions have turns in both portions. Its results and
intervals consequently do not establish generalization to unseen
questions or account for dependence between their turns.

For every fitted probe bank, including HelpSteer2, we select the
regularization penalty by grouped cross-validation on exploration
data, choosing the strongest penalty among exact ties. The rule
and selected penalties appear in \cref{AppCalibrationSensitivity}.

\myparagraph{Probe and rubric measurements}
Each expanded bank has 52 observable properties, 20 reading
contexts or situations, eight evaluator roles, five focus
instructions, and four instructions about the stakes of evaluation.
The wording avoids explicitly naming the benchmarks' rated criteria
or SAJA's criterion dimensions. A property contributes two oriented
response columns; a one-sided instruction contributes one.
Each column is differenced against the baseline in its own call,
centered within context, and screened against that call's five
neutral clauses. The primary fit retains all units whose
exploration bootstrap lower bound for the probe/control RMS ratio
exceeds one. The fixed reference score is from the original bank.
The centered expert target is norm-matched to that reference as in
\eqref{EqnRankingTarget}. Relative ridge penalties range from
$10^{-3}$ to $10^3$ and are selected by five-fold, context-grouped
exploration validation on accuracy.

SAJA's eight broad and 28 granular ratings are elicited together
in one typed-rating call per candidate. Both reported SAJA variants
use RF with 300 trees and seed 42, fitted on exploration data.
The primary probe representation uses three calls per candidate, one for each
question bank; query and question budgets are not matched to SAJA.

\subsection{Data provenance and use}
\label{AppDataProvenance}
We use English-language benchmarks spanning news summarization,
story evaluation, dialogue, instruction following, and reasoning;
mathematical notation and code also occur in JudgeBench.
We reuse released annotations rather than collect new human ratings.
The use considered here is research on automatic evaluation, without
a claim of multilingual or demographic representativeness.
SummEval is loaded from the \texttt{mteb/summeval} mirror at revision
\texttt{bfc1211}. Downloaded benchmark files are retained unchanged
and verified with SHA-256 checksums.

\Cref{TabDataLicenses} reports the notices at the distribution
sources. These do not establish separate rights to all underlying
news articles, writing prompts, or other third-party material.
The USR annotation download does not state a separate data license;
the Apache-2.0 notice for its code is not treated as an annotation
license.

\begin{table}[!htbp]
  \centering
  \small
  \setlength{\tabcolsep}{4pt}
  \begin{tabular}{@{}ll@{}}
    \toprule
    Data source & Reported license\\
    \midrule
    \href{https://huggingface.co/datasets/mteb/summeval}{SummEval mirror} & MIT (dataset card)\\
    \href{https://github.com/dig-team/hanna-benchmark-asg}{HANNA release} & MIT (repository)\\
    \href{https://huggingface.co/datasets/nvidia/HelpSteer2}{HelpSteer2} & CC-BY-4.0\\
    \href{https://huggingface.co/datasets/lmsys/mt_bench_human_judgments}{MT-Bench judgments} & CC-BY-4.0\\
    \href{https://huggingface.co/datasets/ScalerLab/JudgeBench}{JudgeBench} & MIT\\
    \href{https://github.com/alexa/Topical-Chat/blob/master/DATALICENSE}{Topical-Chat base data} & CDLA-Sharing 1.0\\
    \href{https://shikib.com/usr}{USR annotations} & Not specified\\
    \bottomrule
  \end{tabular}
  \caption{Release-level data notices, checked on 13 October 2026.
  Linked sources distinguish the underlying Topical-Chat data from
  the additional USR judgments.}
  \label{TabDataLicenses}
\end{table}

\FloatBarrier
\Needspace{8\baselineskip}
\section{Additional probe-response diagnostics}
\label{AppProbeResponses}
The examples and diagnostics below support the sensitivity
measurements and local interpretation in \cref{SecInterpretation}.

\subsection{Recorded prompt examples}
\label{AppPromptExample}
\Cref{TabAdditionalPromptExamples} changes the attribute while
keeping the same article, summary, Vanilla score, and instruction
template as \cref{TabPromptExample}. The added instruction is
\textquotedblleft When you assign the score, weight \emph{[attribute clause]}
more/less heavily than you normally would.\textquotedblright{}
Each displayed value is one recorded response using this wording,
not an average over the two wordings used for calibration.

\begin{table}[!htbp]
  \centering\small
  \setlength{\tabcolsep}{6pt}
  \begin{tabular}{@{}lp{0.45\textwidth}rrrr@{}}
    \toprule
    Attribute & Attribute clause & $s^+$ & $s^-$ & $\Delta^+$ & $\Delta^-$\\
    \midrule
    Grounding & whether the summary's claims are supported by the source article & $7$ & $6$ & $+1$ & $0$\\[3pt]
    Fluency & the grammar, spelling and word choice of the summary & $7$ & $7$ & $+1$ & $-1$\\[3pt]
    Accuracy & whether the summary's claims are factually correct & $6$ & $6$ & $0$ & $0$\\[3pt]
    \bottomrule
  \end{tabular}
  \caption{Additional recorded Sol examples for the same candidate
  as \cref{TabPromptExample}, with $s^0=6$ throughout.
  Here $s^\pm=s(x;p_k^\pm)$,
  $\Delta^+=s^+-s^0$, and $\Delta^-=s^0-s^-$.
  Grounding changes one side, fluency moves both scores upward,
  and accuracy leaves both unchanged. These illustrate response
  patterns for one input, not average effects of the attributes.}
  \label{TabAdditionalPromptExamples}
\end{table}

\subsection{Repeated scores and neutral instructions}
\label{AppControlVariation}

The repeat reference in \cref{FigAssumptions}(a) uses the unmodified
Vanilla question for each summary in three replicate runs.
Each call bundles this question with other questions; these are not
standalone Vanilla-only calls. We center scores within document and
pool the squared differences over all three run pairs before taking
the square root, without dividing by $\sqrt{2}$. The resulting RMS is
$0.0434$ on the $0$--$4$ scale. Probe-minus-Vanilla displacements are
formed within a call, whereas repeat differences compare calls, so
this ratio is a repeat-variation benchmark, not a decomposition of
signal and noise.

\begin{figure}[htbp]
  \centering
  \begin{minipage}{0.62\linewidth}
    \includegraphics[width=\linewidth]{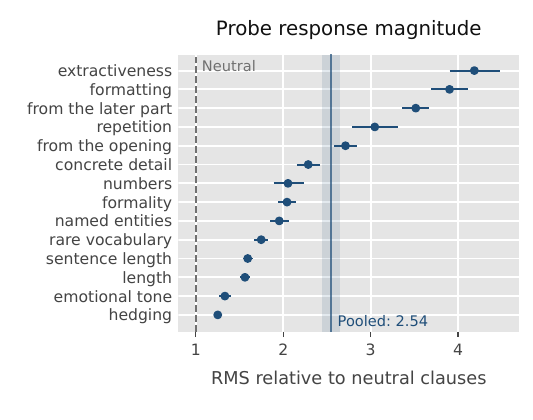}
  \end{minipage}
  \caption{Probe displacements relative to neutral instructions for
  Jev-1.13 on the same 100 SummEval documents as \cref{FigAssumptions}.
  Each point divides a property's RMS displacement, pooled over both
  signs and three runs, by the RMS displacement of five neutral clauses.
  All displacements are centered within document. The dashed line marks
  the neutral reference; the vertical band pools all 14 properties.
  Bars and the band are 95\% document-bootstrap intervals.}
  \label{FigNeutralReference}
\end{figure}

Five neutral clauses, such as ``Take your time and weigh each candidate
fairly,'' provide a second reference. Their RMS displacement from the
same-call Vanilla score is $0.0429$, or $0.988$ times the repeat reference
(95\% interval $[0.955, 1.022]$). Against these neutral clauses, the
pooled probe ratio is $2.54$ ($[2.44, 2.65]$). \Cref{FigNeutralReference}
includes all 14 properties, including extractiveness (reuse of source
wording); ratios range from $1.25$ for hedging to $4.19$ for
extractiveness.

\begin{figure}[!htbp]
  \centering
  \begin{minipage}{0.62\linewidth}
    \includegraphics[width=\linewidth]{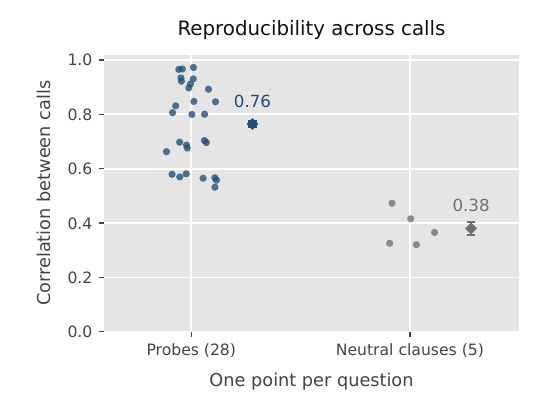}
  \end{minipage}
  \caption{Repeatability of centered displacements on the same three
  Jev runs as \cref{FigAssumptions}. Points are question-level
  correlations; diamonds and bars summarize the mean and its 95\%
  document-bootstrap interval. Vanilla scores have repeat correlation
  $0.997$, which measures score stability, not displacement repeatability.}
  \label{FigRepeatReproducibility}
\end{figure}

The least reproducible probe has correlation $0.53$, above the most
reproducible neutral clause ($0.47$). The neutral mean of $0.38$
indicates that adding a neutral instruction itself has a reproducible
component; the probe mean is $0.76$ (\cref{FigRepeatReproducibility}).

\myparagraph{Variation across judge settings}
For eight quality attributes under single-candidate scoring, the
probe-to-neutral RMS ratio is $1.96$ ($[1.89,2.03]$) for Jev and
$1.15$ ($[1.12,1.18]$) for Sol on the same 100 documents. The
attribute-wise lower bounds exceed one for all eight Jev attributes
and seven of eight Sol attributes. These recorded settings differ in
score elicitation: Jev returns expected ratings, while Sol returns
1--10 integers. They compare probes with neutral instructions;
they do not provide Sol's counterpart to the Vanilla-repeat reference
in \cref{FigAssumptions}(a).

\subsection{Baselines and a local weight interpretation}
\label{AppResponseBaselines}

The wording and joint-instruction experiments use Jev-1.13 on all
100 SummEval documents, with 16 summaries per document.
\Cref{FigProbeResponses} uses ten repeated wording runs; the
supplementary Vanilla-baseline and joint-instruction analyses below
use two runs per design. We report all six recorded properties:
length, concrete detail, extractiveness, repetition, formality,
and hedging. The main plots use an
unchanged-weight control for each attribute and retain the full shared
component. We also report contrasts about the Vanilla baseline used
in~\eqref{EqnResponse}.

The control $p_k^0$ keeps the named attribute explicit while the
requested weight changes. Its within-document-centered displacement from the
Vanilla baseline has $1.21$--$3.07$ times the RMS of the five neutral
clauses, depending on the property, so it cannot be treated as an
unchanged Vanilla baseline. Let
$Z_k(x)=s(x;p_k^0)-s^0(x)$. The shared component about the
Vanilla baseline and the method's oriented response columns satisfy
\begin{align}
  E^{\mathrm{base}}_k
  &= E^{\mathrm{c}}_k+Z_k,
  \notag\\
  \Delta^+_k &= O_k+E^{\mathrm{base}}_k,
  \qquad
  \Delta^-_k = O_k-E^{\mathrm{base}}_k.
  \label{EqnBaselineParts}
\end{align}
Thus, Vanilla-baseline contrasts include the response to adding the
control as well as changes relative to it. Pooled over all six
properties, $E^{\mathrm{base}}$ has $2.25$, $2.26$, and $1.64$ times
the reproducible RMS of $O$ at slight, standard, and strong wording.
The calibration comparison in \cref{SecAblations} retains or
discards this combined component; it does not isolate the predictive
value of $Z$ from that of $E^{\mathrm{c}}$.
The paired representation also fits two coefficients per attribute
rather than one, so this ablation changes both the retained information
and model capacity.

For repetition, the worded control has projection $3.11$ onto the
standard half-difference. Consequently, \emph{slightly more} has
projection $-0.54$ relative to that control but $2.57$ relative to the
Vanilla baseline. The sign of a projected displacement therefore
depends on the reference. Differences between two nonzero wording
responses are unaffected by this change of reference.

For the local interpretation, suppose a smooth function $g(x,w)$
describes the mean score and the two prompts implement small changes
$\pm a_k$ about the weights $w^{\mathrm{c}}_k$ associated
with $p_k^0$. Averaging over judge-call variation, a first-order
expansion gives
\begin{align*}
  \Exs\!\left[(\widehat J^{\mathrm{half}}(X))_k\mid X=x\right]
  \approx \nabla_w g(x,w^{\mathrm{c}}_k)^\top a_k.
\end{align*}
This is a conditional interpretation motivated by local sensitivity
analysis \citep{cook1986local} and perturbation-based surrogate
explanations \citep{ribeiro2016trust}. The attribute-specific controls
need not share a common latent weight vector or reproduce the Vanilla
baseline. A comparison of wordings does not
identify $a_k$ or show that two instructions compose the same weight
changes as they do separately. We consequently report the measured
contrasts without a numerical prediction from smoothness alone.

\subsection{Estimating reproducible response profiles}
\label{AppProbeEstimation}

Let $U_i^{(r)}$ and $V_i^{(r)}$ be two within-document-centered
contrasts for summary $i$ in replicate $r\in\{1,\ldots,R\}$, and let
$n$ be the number of summaries. We average products between
distinct runs:
\begin{align*}
  Q(U,V)
  \defn \frac{1}{nR(R-1)}\sum_{i=1}^n
  \sum_{r\ne r'} U_i^{(r)}V_i^{(r')}.
\end{align*}
With independent, zero-mean noise across runs, the noise contribution
vanishes in expectation. We estimate reproducible RMS by
$\{Q(U,U)\}_+^{1/2}$ and correlation by
$Q(U,V)/\{Q(U,U)Q(V,V)\}^{1/2}$. Pooling over properties averages their
second moments before taking a square root. We resample the 100
documents jointly for all contrasts in 2,000 bootstrap draws.
For \cref{FigProbeResponses}, $R=10$ gives 45 unordered run pairs,
which are averaged rather than treated as independent observations.
All ten runs repeat the same requests on the same documents.
The supplementary analyses use $R=2$.
The main-text shared component is $E^{\mathrm{c}}$ itself, including any component
along $O$.

Each ordinate in \cref{FigProbeResponses}(a) is the projection
coefficient
\[
\frac{Q(D_{k,t},O_{k,\mathrm{standard}})}
{Q(O_{k,\mathrm{standard}},O_{k,\mathrm{standard}})},
\]
where
$D_{k,t}$ is the score under wording $t$ minus the score under
that attribute's control $p_k^0$.
This measures alignment with the standard half-difference, rather than
an absolute score change. Its value at the control is zero by
construction. The categories do not assign numerical steps to wording.

\subsection{Joint instructions}

For each of the 15 pairs of the six recorded properties, we request more or less weight
on both attributes in all four combinations. Writing $s_{uv}$ for the
score with signs $u,v$, the main effect of the first attribute and
the interaction are
\begin{align*}
  M_a &= \frac{s_{++}+s_{+-}-s_{-+}-s_{--}}{4},
 \\
  I_{ab} &= \frac{s_{++}-s_{+-}-s_{-+}+s_{--}}{4}.
\end{align*}
The second main effect $M_b$ exchanges the two attributes. We compare
each main effect with the corresponding single-attribute
half-difference using the projection coefficient defined by $Q$.

\begin{figure}[!htbp]
  \centering
  \begin{minipage}[t]{0.485\linewidth}
    \centering
    \includegraphics[width=\linewidth]{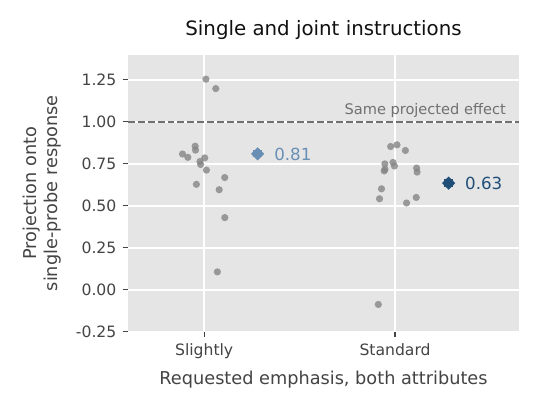}
    \par\smallskip {\small (a)}
  \end{minipage}\hfill
  \begin{minipage}[t]{0.485\linewidth}
    \centering
    \includegraphics[width=\linewidth]{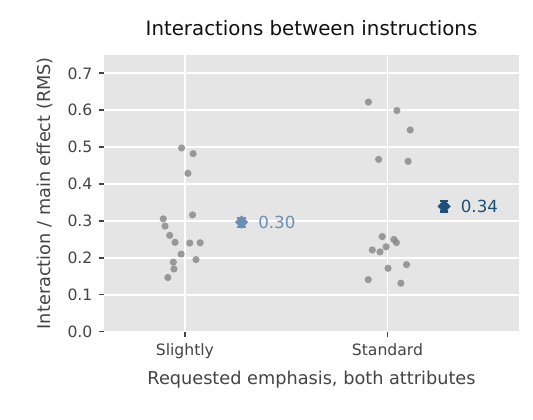}
    \par\smallskip {\small (b)}
  \end{minipage}
  \caption{Joint-instruction diagnostics. Gray points correspond to
  all 15 attribute pairs; diamonds pool their second moments.
  (a)~Projection of joint-instruction main effects onto the
  corresponding single-probe half-differences. The reference at one
  indicates the same projected effect.
  (b)~RMS of the interaction divided by the root mean square of the
  two main effects. Bars on the pooled estimates are 95\%
  document-bootstrap intervals. Neither panel assumes a numerical
  step size for the instructions.}
  \label{FigJointInstructions}
\end{figure}

The pooled projection coefficients are $0.81$ for slight wording and
$0.63$ for standard wording (\cref{FigJointInstructions}(a)). These
coefficients summarize alignment with the single-probe responses;
they do not imply that every response is uniformly rescaled.
The interaction RMS is $0.30$ and $0.34$ of the main-effect RMS,
respectively (\cref{FigJointInstructions}(b)). Thus, the observed
joint instructions are not additive. These measurements concern how
instructions combine; they do not distinguish changes in the
prompt-to-weight map from curvature of a weight-to-score function.

\subsection{Presentation order and evaluation format}
\label{AppPresentationOrder}
\label{AppPairwise}

For the joint-scoring diagnostic in \cref{FigAssumptions}(b,c), Jev
scores all 16 summaries of each document in one call. We use
32 orders: 16 cyclic shifts of a random order, so that every
summary occupies each position once, and 16 shifts of its reversal,
with two additional repeats of one order. Reversing the order gives
a within-document Kendall correlation of $0.57$ for baseline scores,
compared with $0.93$ between repeats. For three chat judges ranking
the summaries, reversal gives $0.68$--$0.72$. The estimated
first-position shift is $-0.16$ on the $0$--$4$ scale. Additive
position effects account for $2.7\%$ of the order-induced variance,
while $93\%$ is an interaction between order and summary.

Subtracting the baseline within the same call does not remove this
sensitivity: probe-response correlation is $0.68$ between repeats
of one order and $0.32$ between different orders. Correlation between
two disjoint sets of averaged orders reaches $0.64$ at four orders
and $0.78$ at eight, against the pointwise reference of $0.70$.
Before calibration, pointwise scoring reaches $0.647$ in mean
Spearman correlation with experts, versus $0.587$ for a single
candidate order. Averaging eight orders reaches $0.647$ and averaging
32 reaches $0.658$. From four orders onward, the paired interval
for the difference from pointwise includes zero; this does not
establish equivalence or a uniform advantage for either format. Averaging a common set of orders for
baseline and probe scores standardizes their treatment without
establishing permutation invariance. Pointwise scoring evaluates
one summary per call and avoids candidate-order effects.

\myparagraph{Pairwise comparisons}

We also compare pairwise judgments with the pointwise reference on the
same 100 SummEval documents. Each summary's score is its mean probability
of winning against the other summaries. Using one presentation order
per pair gives a mean within-document Spearman difference of $-0.077$
from pointwise scoring (95\% interval $[-0.107,-0.050]$). Averaging both
orders reduces the gap to $-0.052$ ($[-0.078,-0.026]$). Both differences
are measured before residual calibration; reversing each pair reduces
the deficit but does not close it in this experiment.

\begin{figure}[htbp]
  \centering
  \begin{minipage}{0.62\linewidth}
    \includegraphics[width=\linewidth]{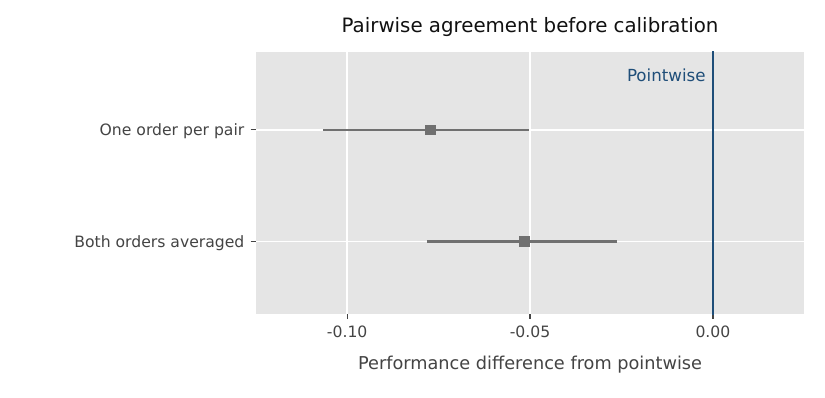}
  \end{minipage}
  \caption{Pairwise judgments relative to pointwise scoring for
  Jev-1.13 on SummEval, before calibration. Each point is the mean
  within-document Spearman difference from the pointwise reference;
  bars are paired 95\% document-bootstrap intervals. The two conditions
  use one presentation order per pair or average both orders.}
  \label{FigPairwise}
\end{figure}

\subsection{Probe diagnostics across tasks}
The original-pool measurements extend the checks in
\cref{SecInterpretation} beyond summarization. All five benchmarks
show probe/control RMS ratios above two and higher repeat
correlation for probes than for neutral clauses. Reversing joint
presentation order changes scores more than repeating the same
order. At the mildest wording level, the reproducible
sign-independent component remains $0.50$--$0.99$ times the
directional component, exceeding the corresponding smooth-response
reference in each benchmark. These observations support measuring
and retaining both response components; they do not by themselves
predict the calibration gains in \cref{TabXBenchStrategies}.

\begin{table}[!htbp]
  \centering
  \small
  \setlength{\tabcolsep}{6pt}
  \renewcommand{\arraystretch}{1.10}
  \begin{tabular}{@{}lcccc@{}}
    \toprule
    Benchmark & RMS ratio & \shortstack{Reproducibility\\probe / neutral} & \shortstack{Kendall\\reverse / repeat} & \shortstack{Even/odd at $\pm1$\\(smooth reference)}\\
    \midrule
    HANNA & $3.12$ & $0.90/0.56$ & $0.57/0.92$ & $0.50\;(0.19)$\\
    Topical-Chat & $2.24$ & $0.85/0.44$ & $0.79/0.95$ & $0.89\;(0.62)$\\
    HelpSteer2 & $2.53$ & $0.88/0.55$ & $0.62/0.97$ & $0.74\;(0.36)$\\
    MT-Bench & $2.48$ & $0.87/0.47$ & $0.72/0.96$ & $0.71\;(0.25)$\\
    JudgeBench & $2.46$ & $0.86/0.53$ & $0.70/0.97$ & $0.99\;(0.24)$\\
    \bottomrule
  \end{tabular}
  \caption{Probe diagnostics on the confirmation split from the 24-unit
  study. The signal ratio pools its 16 properties; reproducibility
  compares independent calls. Kendall compares reversed candidate order
  with an unchanged-order repeat in joint scoring. The final column uses four diagnostic
  properties chosen on exploration data and the unchanged-weight
  instruction as the zero of the wording ladder; reproducible components
  are estimated by between-run cross-products. These are separate
  diagnostic calls, not the 89-unit calibration pool.}
  \label{TabXBenchDiagnostics}
\end{table}

\FloatBarrier
\Needspace{8\baselineskip}
\section{Complete evaluation results}
\label{AppCompleteResults}
\subsection{Full judge comparisons}
\label{AppFullComparisons}
\Cref{TabSummEvalAgreement,TabHannaAgreement} extend the main
comparison to all four judges. The main text focuses on Jev as
a lightweight decision model and Sol as a frontier judge, keeping
the presentation compact amid uneven score-collection reliability.
In particular, Gemma's SummEval G-Eval requires substantial
imputation; Sol also needs fallbacks in SummEval, while its
reported HANNA scores are complete under the response-format
instructions in \cref{AppBaselineAcquisition}.
The acquisition details and missing-score treatment are collected
in \cref{AppBaselineAcquisition}.

\begin{table}[!htbp]
  \centering
  \small
  \setlength{\tabcolsep}{10pt}
  \renewcommand{\arraystretch}{1.1}
  \begin{tabular}{@{}lcccc@{}}
    \toprule
    Method & Jev & Sonnet & Sol & Gemma\\
    \midrule
    \multicolumn{5}{@{}l}{\emph{Rubric-based scoring}}\\[2pt]
    SAJA broad & \shortstack{$0.703$\\[-1pt]{\footnotesize$[0.672,0.732]$}} & \shortstack{$0.657$\\[-1pt]{\footnotesize$[0.624,0.689]$}} & \shortstack{$0.659$\\[-1pt]{\footnotesize$[0.627,0.688]$}} & \shortstack{$0.687$\\[-1pt]{\footnotesize$[0.657,0.716]$}}\\[3pt]
    SAJA granular & \shortstack{$\mathbf{0.745}$\\[-1pt]{\footnotesize$[0.719,0.770]$}} & \shortstack{$0.671$\\[-1pt]{\footnotesize$[0.637,0.705]$}} & \shortstack{$0.707$\\[-1pt]{\footnotesize$[0.679,0.732]$}} & \shortstack{$0.675$\\[-1pt]{\footnotesize$[0.643,0.705]$}}\\[3pt]
    G-Eval & \shortstack{$0.701$\\[-1pt]{\footnotesize$[0.670,0.729]$}} & \shortstack{$\mathbf{0.712}$\\[-1pt]{\footnotesize$[0.683,0.739]$}} & \shortstack{$\mathbf{0.721}$\\[-1pt]{\footnotesize$[0.696,0.745]$}} & \shortstack{$0.682^{\dagger}$\\[-1pt]{\footnotesize$[0.647,0.716]$}}\\[3pt]
    \midrule
    \multicolumn{5}{@{}l}{\emph{Rubric-free scoring}}\\[2pt]
    Vanilla & \shortstack{$0.654$\\[-1pt]{\footnotesize$[0.615,0.691]$}} & \shortstack{$0.645$\\[-1pt]{\footnotesize$[0.607,0.681]$}} & \shortstack{$0.693$\\[-1pt]{\footnotesize$[0.661,0.722]$}} & \shortstack{$0.672$\\[-1pt]{\footnotesize$[0.639,0.708]$}}\\[3pt]
    \rowcolor{gray!10}
    \prism{} (ours) & \shortstack{$0.705$\\[-1pt]{\footnotesize$[0.673,0.735]$}} & \shortstack{$0.684$\\[-1pt]{\footnotesize$[0.647,0.718]$}} & \shortstack{$0.706$\\[-1pt]{\footnotesize$[0.674,0.734]$}} & \shortstack{$\underline{0.695}$\\[-1pt]{\footnotesize$[0.662,0.726]$}}\\[3pt]
    \rowcolor{gray!10}
    \alphaprism{} (ours) & \shortstack{$\underline{0.712}$\\[-1pt]{\footnotesize$[0.683,0.741]$}} & \shortstack{$\underline{0.689}$\\[-1pt]{\footnotesize$[0.653,0.723]$}} & \shortstack{$\underline{0.712}$\\[-1pt]{\footnotesize$[0.682,0.740]$}} & \shortstack{$\mathbf{0.701}$\\[-1pt]{\footnotesize$[0.670,0.730]$}}\\[3pt]
    \bottomrule
  \end{tabular}
  \caption{Full four-judge results on SummEval, measured by mean
  within-document Spearman correlation (higher is better). Brackets give 95\%
  document-bootstrap confidence intervals. Bold and underline mark
  the best and second-best displayed means, including ties.
  Groups distinguish scoring protocols, not probe provenance;
  \prism{} fixes $\alpha=1$ and \alphaprism{} learns it.
  $\dagger$ Gemma's G-Eval includes midpoint imputation for 33.3\%
  of criterion scores (\cref{AppGEval}).}
  \label{TabSummEvalAgreement}
\end{table}

\begin{table}[!htbp]
  \centering\small

  \setlength{\tabcolsep}{10pt}
  \begin{tabular}{@{}lcccc@{}}
    \toprule
    Method & Jev & Sonnet$^{\dagger}$ & Sol & Gemma$^{\dagger}$\\
    \midrule
    \multicolumn{5}{@{}l}{\emph{Rubric-based scoring}}\\[2pt]
    SAJA broad & \shortstack{$0.524$\\[-1pt]{\footnotesize$[0.484,0.565]$}} & \shortstack{$0.389$\\[-1pt]{\footnotesize$[0.333,0.445]$}} & \shortstack{$0.443$\\[-1pt]{\footnotesize$[0.390,0.492]$}} & \shortstack{$0.364$\\[-1pt]{\footnotesize$[0.309,0.414]$}}\\[3pt]
    SAJA granular & \shortstack{$0.549$\\[-1pt]{\footnotesize$[0.508,0.590]$}} & \shortstack{$\mathbf{0.500}$\\[-1pt]{\footnotesize$[0.450,0.548]$}} & \shortstack{$0.528$\\[-1pt]{\footnotesize$[0.482,0.574]$}} & \shortstack{$0.466$\\[-1pt]{\footnotesize$[0.410,0.517]$}}\\[3pt]
    G-Eval & \shortstack{$0.528$\\[-1pt]{\footnotesize$[0.483,0.575]$}} & \shortstack{$\underline{0.497}$\\[-1pt]{\footnotesize$[0.452,0.543]$}} & \shortstack{$\mathbf{0.587}$\\[-1pt]{\footnotesize$[0.543,0.632]$}} & \shortstack{$\mathbf{0.497}$\\[-1pt]{\footnotesize$[0.449,0.544]$}}\\[3pt]
    \midrule
    \multicolumn{5}{@{}l}{\emph{Rubric-free scoring}}\\[2pt]
    Vanilla & \shortstack{$0.459$\\[-1pt]{\footnotesize$[0.409,0.505]$}} & \shortstack{$0.432$\\[-1pt]{\footnotesize$[0.384,0.478]$}} & \shortstack{$0.462$\\[-1pt]{\footnotesize$[0.428,0.497]$}} & \shortstack{$0.437$\\[-1pt]{\footnotesize$[0.398,0.474]$}}\\[3pt]
    \rowcolor{gray!10}
    \prism{} (ours) & \shortstack{$\underline{0.599}$\\[-1pt]{\footnotesize$[0.558,0.639]$}} & \shortstack{$0.483$\\[-1pt]{\footnotesize$[0.434,0.531]$}} & \shortstack{$0.546$\\[-1pt]{\footnotesize$[0.502,0.588]$}} & \shortstack{$\underline{0.475}$\\[-1pt]{\footnotesize$[0.427,0.521]$}}\\[3pt]
    \rowcolor{gray!10}
    \alphaprism{} (ours) & \shortstack{$\mathbf{0.603}$\\[-1pt]{\footnotesize$[0.562,0.642]$}} & \shortstack{$0.484$\\[-1pt]{\footnotesize$[0.432,0.531]$}} & \shortstack{$\underline{0.557}$\\[-1pt]{\footnotesize$[0.514,0.598]$}} & \shortstack{$0.473$\\[-1pt]{\footnotesize$[0.424,0.519]$}}\\[3pt]
    \bottomrule
  \end{tabular}
  \caption{Full four-judge results on HANNA: mean within-prompt Spearman
  correlation (higher is better), with 95\% prompt-bootstrap intervals.
  All methods use the same 96 prompts; fitted methods share five outer
  folds. SAJA uses 1--10 scores. Vanilla/probe scores use 0--100 for
  Sonnet, 1--10 for Sol and Gemma, and expected ratings for Jev.
  G-Eval uses six HANNA criteria with model-specific response-format
  adjustments (\cref{AppBaselineAcquisition}). Bold and underline mark the
  best and second-best displayed means.
  $\dagger$ Sonnet and Gemma include mean-imputed failed scores.}
  \label{TabHannaAgreement}
\end{table}

\subsection{Additional benchmark results}
\label{AppAdditionalResults}
\Cref{TabXBenchContrasts} reports paired accuracy differences for
the primary fits in \cref{TabBeyondSummEval}, plus the separate
HANNA confirmation study. Splits and acquisition are specified
in \cref{AppCrossBenchmark}; probe-pool ablations appear in
\cref{AppExpandedPool}. These intervals condition on the fitted
predictor and selected bank.

\begin{table}[!htbp]
  \centering
  \small
  \setlength{\tabcolsep}{5pt}
  \renewcommand{\arraystretch}{1.10}
  \begin{tabular}{@{}lcc@{}}
    \toprule
    Benchmark & $\Delta$ Vanilla & $\Delta$ SAJA broad\\
    \midrule
    HANNA & $+0.063\;[+0.041,+0.085]$ & $+0.041\;[+0.019,+0.062]$\\
    Topical-Chat & $+0.002\;[+0.000,+0.005]$ & $+0.038\;[-0.003,+0.079]$\\
    HelpSteer2 & $+0.016\;[-0.003,+0.040]$ & $+0.090\;[+0.027,+0.154]$\\
    MT-Bench & $+0.010\;[-0.011,+0.033]$ & $+0.021\;[-0.013,+0.059]$\\
    JudgeBench & $+0.002\;[-0.003,+0.009]$ & $+0.115\;[+0.057,+0.171]$\\
    \bottomrule
  \end{tabular}
  \caption{Primary \prism{} minus Vanilla and SAJA broad, with paired 95\% context-bootstrap intervals. The 2,000 resamples hold fitted models and selected banks fixed; intervals are unadjusted for the family of reported comparisons. Topical-Chat's interval against the Vanilla has an exact lower endpoint of zero.}
  \label{TabXBenchContrasts}
\end{table}

\FloatBarrier
\Needspace{8\baselineskip}
\section{Additional ablations and correction profiles}
\label{AppAdditionalAblations}
This section specifies the representation, regularization, and
probe-bank comparisons supporting \cref{SecAblations}, followed by
the complete correction profiles and supplementary diagnostics.

\subsection{Response representations and diagonal Taylor calibration}
\label{AppResponseAblation}
The three predictors in \cref{TabCalibrationAblations} use the same
recorded SummEval scores, targets, and document folds, fixing the
Vanilla coefficient at one. For each attribute, define
\begin{align*}
  O_k(x)&=\frac{s(x;p_k^+)-s(x;p_k^-)}{2}
         =\frac{\Delta_k^+(x)+\Delta_k^-(x)}{2},\\
  H_k(x)&=s(x;p_k^+)+s(x;p_k^-)-2s^0(x)\\
        &=\Delta_k^+(x)-\Delta_k^-(x).
\end{align*}
The half-difference predictor uses only $O$ and fits one coefficient
per attribute. Paired \prism{} uses both oriented responses and fits
two coefficients per attribute.

\myparagraph{Diagonal Taylor predictor}
To make the second-order comparison explicit, consider the quadratic
surrogate in instruction coordinates $u\in\real^K$,
\begin{align*}
  \widetilde g_x(u)
  =s^0(x)+\sum_k O_k(x)u_k+\frac12\sum_k H_k(x)u_k^2.
\end{align*}
It matches the recorded scores at $u=0$ and $u=\pm e_k$, where
$e_k$ is the $k$th coordinate unit vector.
If these coordinates were symmetric unit steps of a smooth score
function, $O_k$ and $H_k$ would be finite-difference surrogates for
its gradient and Hessian diagonal. The actual prompt wordings do
not identify such numerical steps: this is an explicit quadratic
model of the observed contrasts, without cross-attribute Hessian
terms or a claim to recover the judge's internal derivatives.
Evaluating it at a learned shared displacement $u=-\delta$ gives
\begin{align*}
  \yhat_{\mathrm{Taylor}}(x)
  =s^0(x)-\sum_k O_k(x)\widehat\delta_k
         +\frac12\sum_k H_k(x)\widehat\delta_k^2.
\end{align*}

\myparagraph{Fitting}
Let $\mathbf v_T$ be the centered Vanilla scores on training
contexts, $\mathbf O_T,\mathbf H_T$ their centered contrast matrices,
and $\mathbf y_T^*$ the targets in \eqref{EqnRankingTarget}. Writing
$\delta\odot\delta$ for the coordinatewise square, we fit
\begin{align*}
  \mathbf f_T(\delta)
    &=\mathbf v_T-\mathbf O_T\delta
      +\tfrac12\mathbf H_T(\delta\odot\delta),\\
  \widehat\delta&\in\argmin_{\delta\in\real^K}
    \Big\{\|\mathbf f_T(\delta)-\mathbf y_T^*\|_2^2\\
    &\qquad\qquad
      +\gamma\frac{\|\mathbf O_T\|_F^2}{K}\|\delta\|_2^2\Big\}.
\end{align*}
The penalty grid and nested validation rule are the same as in
\cref{AppCalibrationSensitivity}. Nonlinear least squares starts
from the corresponding linear ridge fit, zero, and the negated
linear fit; we retain the solution with lowest penalized training
loss. The recorded solver uses an analytic Jacobian, at most 1,000
evaluations, and $10^{-10}$ convergence tolerances. For held-out
predictions, the centered correction is applied to the original
Vanilla score, preserving its context mean.

\myparagraph{Relation to paired \prism{}}
Since $(O,H)$ is an invertible transformation of
$(\Delta^+,\Delta^-)$, the Taylor predictor adds no measurements.
Its coefficient relation can be seen by writing the paired score as
\begin{align*}
  \yhat
  &=s^0-\sum_k O_k(\beta_k^++\beta_k^-)\\
  &\quad-\tfrac12\sum_k H_k(\beta_k^+-\beta_k^-).
\end{align*}
The Taylor form requires
$\beta_k^++\beta_k^-=\delta_k$ and
$\beta_k^+-\beta_k^-=-\delta_k^2$; paired \prism{} learns both
combinations freely. Thus, the ablation tests a constrained
quadratic correction rather than adding Hessian information to
the surrogate Jacobian. The paired 95\% intervals for paired
\prism{} minus the half-difference are $[-0.002,0.020]$ for Jev and
$[-0.012,0.009]$ for Sol. The intervals for Taylor minus paired
\prism{} are $[-0.005,0.006]$ for Jev, $[-0.006,0.008]$ for
Sonnet, $[-0.008,0.009]$ for Sol, and $[-0.001,0.012]$ for Gemma.
All three predictors reuse the same saved responses.

\subsection{Regularization and the baseline coefficient}
\label{AppCalibrationSensitivity}

We parameterize the ridge penalty in~\eqref{EqnRidge} as
$\lambda=\gamma\|\mathbf J_T\|_F^2/(n_Tq)$, where the centered
training matrix $\mathbf J_T$ has $n_T$ rows and $q$ columns.
Thus, $\gamma$ measures penalty strength relative to the mean squared
training response. With baseline--human residual vector $\mathbf d_T$,
the equivalent normal equations and search grid are
\begin{gather}
  \left(\mathbf J_T^\top\mathbf J_T+
  \frac{\gamma\|\mathbf J_T\|_F^2}{q}I_q\right)\widehat\beta
  =\mathbf J_T^\top\mathbf d_T,
  \notag\\
  \gamma\in\{10^{-3},10^{-2},10^{-1},1,10,100,1000\}.
  \label{EqnRelativePenalty}
\end{gather}
Columns are not individually standardized. \Cref{TabRidgeSensitivity} fixes
$\gamma$ across folds; the selected row instead chooses it
separately using inner validation, with the largest penalty chosen
among exact CV-score ties. The fixed-grid results are a
sensitivity analysis and are not used to choose the main result.

\begin{table}[!htbp]
  \centering
  \small
  \setlength{\tabcolsep}{13pt}
  \begin{tabular}{@{}lrrrr@{}}
    \toprule
    Relative penalty $\gamma$ & Jev & Sonnet & Sol & Gemma\\
    \midrule
    $10^{-3}$ & $0.7001$ & $0.6842$ & $0.7084$ & $0.6952$\\
    $10^{-2}$ & $0.6996$ & $0.6849$ & $0.7079$ & $0.6943$\\
    $10^{-1}$ & $0.7047$ & $0.6846$ & $0.7081$ & $0.6949$\\
    $1$ & $0.7016$ & $0.6822$ & $0.7062$ & $0.6922$\\
    $10$ & $0.6768$ & $0.6561$ & $0.6984$ & $0.6810$\\
    $100$ & $0.6575$ & $0.6479$ & $0.6952$ & $0.6752$\\
    $1000$ & $0.6548$ & $0.6479$ & $0.6948$ & $0.6753$\\
    \midrule
    Inner-validation selection & $0.7047$ & $0.6840$ & $0.7057$ & $0.6952$\\
    Vanilla & $0.6543$ & $0.6448$ & $0.6926$ & $0.6722$\\
    \bottomrule
  \end{tabular}
  \caption{Regularization sensitivity of paired-response calibration:
  mean within-document Spearman correlation on the same outer folds.
  A finite, strongly shrunk correction can still break baseline
  ties, so its rank correlation need not equal the baseline's.}
  \label{TabRidgeSensitivity}
\end{table}

Jev selects $\gamma=0.1$ in all five folds. Sonnet selects
$0.1$ in four and $1$ in one; Sol selects $0.1$ in three and
$1$ in two. Gemma selects $10^{-3}$ in three, $10^{-2}$ in
one, and $10^{-1}$ in one. For \alphaprism{},
we minimize the same centered squared loss with prediction
$\alpha C s^0-C\widehat J\beta$, leaving $\alpha$
unpenalized and penalizing $\beta$ by~\eqref{EqnRelativePenalty}.
The fitted $\alpha$ ranges across folds are $0.655$--$0.699$
for Jev, $0.714$--$0.743$ for Sonnet, $0.787$--$0.814$ for Sol,
and $0.734$--$0.752$ for Gemma. Each outer fold reselects
$\gamma$ for \alphaprism{} on the same inner folds
and grid.
Learning $\alpha$ increases mean Spearman by $0.005$--$0.008$
across the four SummEval settings. Sol's paired gain is $0.0065$
with 95\% interval $[0.0010,0.0127]$; the other three intervals
include zero.

\myparagraph{Exact validation ties}
For each additional benchmark, cross-validation selects the largest
penalty among exact maximizers of exploration-set accuracy,
without rounding, tolerance, or an evaluation-score tie breaker.
\Cref{TabXBenchTiePolicy} lists the maximizing penalties and the
selected value for each primary expanded-bank fit. SummEval's
four primary fits have unique maximizers.

\begin{table}[!htbp]
  \centering
  \small
  \setlength{\tabcolsep}{7pt}
  \renewcommand{\arraystretch}{1.10}
  \begin{tabular}{@{}llr@{}}
    \toprule
    Benchmark & CV maximizers $\gamma$ & Selected $\gamma$\\
    \midrule
    HANNA & $\{1\}$ & $1$\\
    Topical-Chat & $\{1000\}$ & $1000$\\
    HelpSteer2 & $\{0.01,0.1,10,100\}$ & $100$\\
    MT-Bench & $\{1\}$ & $1$\\
    JudgeBench & $\{100,1000\}$ & $1000$\\
    \bottomrule
  \end{tabular}
  \caption{Regularization selected on exploration data for each primary
  expanded-bank fit. The strongest penalty among exact CV-score
  maximizers is selected; confirmation scores do not enter this selection.}
  \label{TabXBenchTiePolicy}
\end{table}

\subsection{Robustness to probe attributes}
\label{AppProbeSelection}
\myparagraph{Quality versus style attributes}

\Cref{TabProbeBankDefinitions} lists the complete two banks.
For each attribute clause, the positive templates are
\textquotedblleft When you assign the score, weight \emph{[clause]}
more heavily than you normally would\textquotedblright{} and
\textquotedblleft Let \emph{[clause]} count for more than usual in
the score you give\textquotedblright{}; the negative templates
replace \emph{more} with \emph{less}. The combined bank is their
16-attribute union. Each bank contributes 16 response columns
after averaging its two wordings per sign; the union contributes 32.

\begin{table}[!htbp]
  \centering\small
  \setlength{\tabcolsep}{7pt}
  \begin{tabular}{@{}lp{0.76\textwidth}@{}}
    \toprule
    Attribute & Exact attribute clause inserted in the templates\\
    \midrule
    \multicolumn{2}{@{}l}{\emph{Quality bank (eight attributes)}}\\[2pt]
    Accuracy & whether the summary's claims are factually correct\\[2pt]
    Grounding & whether the summary's claims are supported by the source article\\[2pt]
    Coverage & how much of the article's important content the summary includes\\[2pt]
    Coherence & how well-organised the summary is and how it flows\\[2pt]
    Fluency & the grammar, spelling and word choice of the summary\\[2pt]
    Engagement & how engaging the summary is to read\\[2pt]
    Helpfulness & how well the summary would serve a reader in place of the article\\[2pt]
    Safety & whether the summary is free of harmful or inappropriate content\\[2pt]
    \midrule
    \multicolumn{2}{@{}l}{\emph{Style bank (eight attributes)}}\\[2pt]
    Length & how long the summary is\\[2pt]
    Specificity & how many concrete names, numbers and figures the summary cites\\[2pt]
    Lead & how closely the summary follows the article's opening\\[2pt]
    Polish & how polished and eloquent the writing is\\[2pt]
    Repetition & how much the summary restates its own points\\[2pt]
    Confidence & how assertive and unhedged the summary's wording is\\[2pt]
    Extractiveness & how much the summary reuses the article's own wording\\[2pt]
    Structure & how much visible structure the summary uses, such as lists or headings\\[2pt]
    \bottomrule
  \end{tabular}
  \caption{Complete probe banks in the matched Jev--SummEval
  robustness comparison of \cref{TabProbeAttributes}. Quality
  includes rubric-derived criteria; Style uses eight distinct
  named properties without those explicit criteria. Distinct names
  do not imply that style is statistically independent of quality.
  This eight-attribute style bank differs from the seven-attribute
  Sonnet/Gemma bank in \cref{AppScoreAcquisition}.}
  \label{TabProbeBankDefinitions}
\end{table}

The comparison tests whether calibration needs probes that reproduce
the rubric's criteria. The similarly performing style bank supplies
correction signals without explicitly asking about those criteria;
its union with the quality bank provides little additional gain.

The quality run's Vanilla scores are fixed in all three predictors.
The response differences use each bank's own
recorded baseline and are centered within documents. We refit
\prism{} for each bank and their union using the same five outer and
five inner document folds, penalty grid, and tie rule. The paired
style-minus-quality gain is $0.005$ with 95\% interval
$[-0.008,0.018]$; the union gains $0.002$ over style
$[-0.005,0.009]$ and $0.007$ over quality $[-0.004,0.018]$.
Intervals use 2,000 bootstrap samples of the same 100 documents.
These small observed differences support robustness in this matched
setting; they are not a formal equivalence test.

\myparagraph{Attribute subsets}
For probe selection, we enumerate every nonempty subset of the
recorded attributes: 255 subsets each for Jev and Sol, and 127
each for Sonnet and Gemma. Each retained attribute keeps both
oriented response columns. Each subset repeats the nested ridge
selection on the original document folds. No subset is chosen
using its outer-fold score. These are feature-removal comparisons
on already collected responses, not new judge calls using shortened
prompt banks. Each point in \cref{FigCorrectionProfiles}(c) is the
median over all $\binom{K}{k}$ subsets of size $k$ from a bank of
$K$ attributes; bands show the 10th--90th percentiles across these
subsets, not confidence intervals over documents.

\begin{table}[!htbp]
  \centering
  \small
  \setlength{\tabcolsep}{9pt}
  \begin{tabular}{@{}lcccc@{}}
    \toprule
    Judge & Full bank & After one omission & Four-probe gain & Above baseline\\
    \midrule
    Jev & $0.705$ & $[0.687,0.706]$ & $+0.043$ & $70/70$\\
    Sonnet & $0.684$ & $[0.678,0.686]$ & $+0.036$ & $35/35$\\
    Sol & $0.706$ & $[0.704,0.710]$ & $+0.011$ & $68/70$\\
    Gemma & $0.695$ & $[0.691,0.697]$ & $+0.020$ & $35/35$\\
    \bottomrule
  \end{tabular}
  \caption{Probe-selection robustness. Full-bank and omission values
  are mean within-document Spearman correlations; the omission
  brackets give the full range across removed attributes. The
  four-probe gain is the median across all four-attribute subsets.
  The last column counts subsets whose mean exceeds the Vanilla
  judge, without asserting significance for every subset.}
  \label{TabProbeSelection}
\end{table}

\subsection{Full correction profiles}
\label{AppCorrectionProfiles}

\begin{figure}[!htbp]
  \centering
  \begin{minipage}{0.60\linewidth}
    \includegraphics[width=\linewidth]{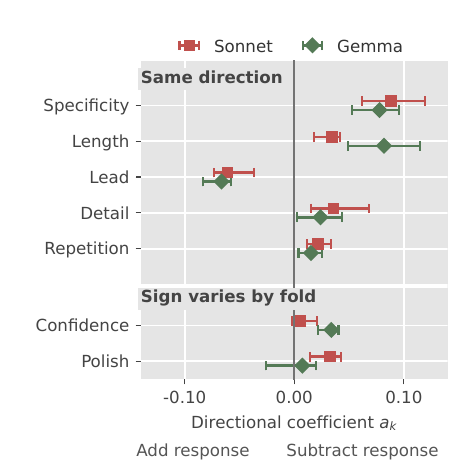}
  \end{minipage}
  \caption{Directional corrections for Sonnet and Gemma. Points are
  mean normalized coefficients $a_k$; bars span five overlapping folds.
  Positive coefficients subtract the response, negative ones add it.
  This bank and elicitation format differ from Jev/Sol in
  \cref{FigCorrectionProfiles}(a); see \cref{TabProbeBankCoverage}.
  A variable sign does not imply no effect.}
  \label{FigStyleCorrectionProfiles}
\end{figure}

For each attribute, the paired correction has the exact decomposition
\begin{align*}
  \Delta^+_k\widehat\beta^+_k
  +\Delta^-_k\widehat\beta^-_k
  &=O_k(\widehat\beta^+_k+\widehat\beta^-_k)\\
  &\quad+E^{\mathrm{base}}_k(\widehat\beta^+_k-\widehat\beta^-_k).
\end{align*}
The shared response is defined about the Vanilla baseline, as in
\eqref{EqnBaselineParts}. For training fold $T$, define
\begin{align*}
  a_k^{(T)}
  &\defn(\widehat\beta^+_k+\widehat\beta^-_k)
  \frac{\operatorname{RMS}_T(CO_k)}
       {\operatorname{RMS}_T(C s^0)},
 \\
  e_k^{(T)}
  &\defn(\widehat\beta^+_k-\widehat\beta^-_k)
  \frac{\operatorname{RMS}_T(CE^{\mathrm{base}}_k)}
       {\operatorname{RMS}_T(C s^0)},
\end{align*}
where $C$ centers each document and RMS pools its training summaries.
\Cref{TabFullCorrectionProfiles} reports fold means and ranges.
This normalization describes the size of each fitted component;
it does not remove dependence on correlated columns, the probe
bank, or the ridge penalty. In particular, the entries are not
additive fractions of explained variance. Stability under
rewording or independent repeated calls is distinct from the
calibration-fold variation reported here.

Across the four judges, 24 of 30 directional coefficients retain
their sign in all five folds. These approximate bias coordinates
depend on the recorded bank and elicitation protocol; model-only
comparisons require matching both (\cref{TabProbeBankCoverage}).

\begin{table}[!htbp]
  \centering
  \small
  \setlength{\tabcolsep}{8pt}
  \begin{tabular}{@{}llrr@{}}
    \toprule
    Judge & Attribute & Directional $a_k$ & Shared $e_k$\\
    \midrule
    Jev & Accuracy & $-0.095\;[-0.122,-0.082]$ & $0.004\;[-0.007,0.016]$\\
     & Coherence & $-0.003\;[-0.018,0.004]$ & $0.003\;[-0.015,0.021]$\\
     & Coverage & $0.062\;[0.039,0.078]$ & $-0.018\;[-0.024,-0.008]$\\
     & Engagement & $-0.042\;[-0.054,-0.033]$ & $-0.096\;[-0.119,-0.073]$\\
     & Fluency & $-0.087\;[-0.105,-0.055]$ & $-0.144\;[-0.151,-0.134]$\\
     & Grounding & $-0.021\;[-0.029,-0.013]$ & $0.046\;[0.039,0.054]$\\
     & Helpfulness & $0.021\;[0.012,0.030]$ & $-0.077\;[-0.117,-0.050]$\\
     & Safety & $-0.006\;[-0.013,0.001]$ & $0.035\;[0.032,0.040]$\\
    \midrule
    Sonnet & Confidence & $0.005\;[-0.002,0.021]$ & $-0.011\;[-0.018,0.001]$\\
     & Detail & $0.036\;[0.015,0.068]$ & $-0.019\;[-0.033,-0.011]$\\
     & Lead & $-0.061\;[-0.074,-0.037]$ & $-0.025\;[-0.047,-0.009]$\\
     & Length & $0.035\;[0.018,0.042]$ & $-0.071\;[-0.088,-0.058]$\\
     & Polish & $0.033\;[0.014,0.042]$ & $0.003\;[-0.012,0.020]$\\
     & Repetition & $0.021\;[0.012,0.033]$ & $-0.022\;[-0.025,-0.018]$\\
     & Specificity & $0.088\;[0.062,0.119]$ & $-0.097\;[-0.116,-0.061]$\\
    \midrule
    Sol & Accuracy & $0.040\;[0.026,0.061]$ & $-0.003\;[-0.012,0.004]$\\
     & Coherence & $0.008\;[-0.003,0.021]$ & $-0.028\;[-0.047,-0.006]$\\
     & Coverage & $0.031\;[0.017,0.046]$ & $-0.049\;[-0.084,-0.026]$\\
     & Engagement & $-0.007\;[-0.014,0.000]$ & $-0.003\;[-0.018,0.021]$\\
     & Fluency & $-0.059\;[-0.069,-0.042]$ & $-0.074\;[-0.095,-0.058]$\\
     & Grounding & $0.035\;[0.017,0.049]$ & $-0.058\;[-0.082,-0.033]$\\
     & Helpfulness & $0.014\;[0.004,0.028]$ & $0.051\;[0.020,0.076]$\\
     & Safety & $0.001\;[-0.003,0.005]$ & $-0.010\;[-0.015,-0.003]$\\
    \midrule
    Gemma & Confidence & $0.034\;[0.021,0.040]$ & $0.029\;[0.008,0.059]$\\
     & Detail & $0.024\;[0.002,0.044]$ & $0.078\;[0.042,0.093]$\\
     & Lead & $-0.066\;[-0.083,-0.058]$ & $-0.047\;[-0.077,-0.032]$\\
     & Length & $0.082\;[0.049,0.115]$ & $-0.042\;[-0.068,-0.005]$\\
     & Polish & $0.007\;[-0.026,0.020]$ & $-0.059\;[-0.082,-0.040]$\\
     & Repetition & $0.015\;[0.004,0.025]$ & $-0.101\;[-0.114,-0.091]$\\
     & Specificity & $0.078\;[0.053,0.096]$ & $-0.116\;[-0.125,-0.104]$\\
    \bottomrule
  \end{tabular}
  \caption{Full profiles of the paired-response correction. Each entry is
  the mean over five training folds followed by the minimum and maximum
  in brackets; these ranges are not confidence intervals. Positive
  contributions are subtracted in the deployed score.}
  \label{TabFullCorrectionProfiles}
\end{table}

\FloatBarrier
\subsection{Expanding and selecting the probe pool}
\label{AppExpandedPool}
\Cref{TabXBenchStrategies} compares the recorded probe pools under
the common protocol in \cref{AppCrossBenchmark}. More probes do not
consistently improve calibration, even after screening for reliable
score movement. Detectable responses therefore need not supply
additional predictive information.

\begin{table}[!htbp]
  \centering
  \small
  \setlength{\tabcolsep}{4pt}
  \renewcommand{\arraystretch}{1.10}
  \begin{tabular}{@{}lccccc@{}}
    \toprule
    Method & HANNA & Topical-Chat & HelpSteer2 & MT-Bench & JudgeBench\\
    \midrule
    \multicolumn{6}{@{}l}{\emph{Rubric-based}}\\[2pt]
    SAJA broad & $0.682$ & $0.799$ & $0.665$ & $0.808$ & $0.676$\\
    SAJA granular & $0.695$ & $0.831$ & $0.710$ & $0.805$ & $0.707$\\
    \midrule
    \multicolumn{6}{@{}l}{\emph{Rubric-free}}\\[2pt]
    Vanilla & $0.659$ & $0.835$ & $0.739$ & $0.818$ & $0.789$\\
    \prism{}: all passing (primary) & $0.723$ & $0.837$ & $0.755$ & $0.828$ & $0.791$\\
    Original pool: passing units & $0.726$ & $0.835$ & $0.766$ & $0.820$ & $0.791$\\
    Added pool: passing units & $0.703$ & $0.827$ & $0.750$ & $0.831$ & $0.791$\\
    Expanded pool: forward selection & $0.726$ & $0.812$ & $0.729$ & $0.831$ & $0.780$\\
    Expanded pool: all units & $0.718$ & $0.837$ & $0.755$ & $0.831$ & $0.791$\\
    Properties only & $0.720$ & $0.837$ & $0.761$ & $0.834$ & $0.791$\\
    One-sided instructions only & $0.700$ & $0.834$ & $0.761$ & $0.811$ & $0.791$\\
    \bottomrule
  \end{tabular}
  \caption{Probe-pool strategy comparisons (accuracy $\uparrow$).
  Every fitted predictor uses exploration labels only, with the
  penalty selected separately by grouped cross-validation and exact
  ties favoring stronger regularization. The primary bank retains
  all signal-screened units.}
  \label{TabXBenchStrategies}
\end{table}

The signal screen retains 74, 82, 85, 79, and 84 units for HANNA,
Topical-Chat, HelpSteer2, MT-Bench, and JudgeBench. Secondary forward
selection adds the largest exploration nested-CV gain, stopping
below $0.001$ or at 20 units; it selects 3, 1, 3, 3, and 3 units.
The original-pool rows apply the same screen and regularization
selection to 24 units.
For the expanded primary HANNA fit, within-context Spearman changes
from $0.414$ to $0.568$.

All 260 property units tested across the five expanded banks pass
the neutral-control screen, but only 100 individually improve
exploration nested-CV accuracy by more than $0.002$. This exploratory
count distinguishes measurable response from predictive usefulness;
it is not a held-out test of each unit.

\subsection{Score resolution}
\label{AppScoreResolution}
For Sonnet on HANNA, changing the scale from 1--10 to 0--100 reduces
the tied-pair share from
55.7\% to 19.4\%, and zero probe responses from 80.6\% to 55.7\%.
Spearman changes from $0.445$ to $0.432$ for Vanilla, $0.430$ to
$0.483$ for \prism{}, and $0.439$ to $0.484$ for \alphaprism{}.
The paired \prism{} gain across scales is $0.054$ $[0.019,0.090]$;
its 0--100 gain over Vanilla is $0.052$ $[0.021,0.082]$.
Finer scores preserve more usable variation, although the changed
anchors and missing-response rates also belong to this comparison.
The model, bank, and folds are fixed as described in
\cref{AppScoreAcquisition}.


\begin{thebibliography}{CWSGM24}

\bibitem[Alm26]{almeida2026jev}
Diogo Almeida.
\newblock \href{https://typesafe.ai/blog/introducing-system-one-models-and-jev}{Introducing {System One Models} \& {Jev}.}
\newblock {TypeSafe AI} blog, 2026.

\bibitem[CCSC22]{chhun2022hanna}
Cyril Chhun, Pierre Colombo, Fabian~M. Suchanek, and Chlo{\'e} Clavel.
\newblock \href{https://aclanthology.org/2022.coling-1.509/}{Of human criteria and automatic metrics: A benchmark of the
  evaluation of story generation.}
\newblock In Nicoletta Calzolari, Chu-Ren Huang, Hansaem Kim, James
  Pustejovsky, Leo Wanner, Key-Sun Choi, Pum-Mo Ryu, Hsin-Hsi Chen, Lucia
  Donatelli, Heng Ji, Sadao Kurohashi, Patrizia Paggio, Nianwen Xue, Seokhwan
  Kim, Younggyun Hahm, Zhong He, Tony~Kyungil Lee, Enrico Santus, Francis Bond,
  and Seung-Hoon Na, editors, {\em Proceedings of the 29th International
  Conference on Computational Linguistics}, pages 5794--5836, Gyeongju,
  Republic of Korea, October 2022. International Committee on Computational
  Linguistics.

\bibitem[Coo86]{cook1986local}
R.~Dennis Cook.
\newblock \href{http://dx.doi.org/10.1111/j.2517-6161.1986.tb01398.x}{Assessment of local influence.}
\newblock {\em Journal of the Royal Statistical Society: Series B
  (Methodological)}, 48(2):133--155, January 1986.

\bibitem[CWSGM24]{cohenwang2024contextcite}
Benjamin Cohen-Wang, Harshay Shah, Kristian Georgiev, and Aleksander M\k{a}dry.
\newblock \href{https://proceedings.neurips.cc/paper_files/paper/2024/file/adbea136219b64db96a9941e4249a857-Paper-Conference.pdf}{{ContextCite}: Attributing model generation to context.}
\newblock In A.~Globerson, L.~Mackey, D.~Belgrave, A.~Fan, U.~Paquet,
  J.~Tomczak, and C.~Zhang, editors, {\em Advances in Neural Information
  Processing Systems}, volume~37, pages 95764--95807. Curran Associates, Inc.,
  2024.

\bibitem[DSS26]{deusser2026jev}
Tobias Deu{\ss}er, Lorenz Sparrenberg, and Rafet Sifa.
\newblock \href{https://arxiv.org/abs/2609.37647}{Evaluating and benchmarking the {System One Model Jev}, 2026.}
\newblock arXiv:2609.37647.

\bibitem[FKM{\etalchar{+}}21]{fabbri2021summeval}
Alexander~R. Fabbri, Wojciech Kry{\'s}ci{\'n}ski, Bryan McCann, Caiming Xiong,
  Richard Socher, and Dragomir Radev.
\newblock \href{https://aclanthology.org/2021.tacl-1.24/}{{S}umm{E}val: Re-evaluating summarization evaluation.}
\newblock {\em Transactions of the Association for Computational Linguistics},
  9:391--409, 2021.

\bibitem[GSZ{\etalchar{+}}25]{gupta2025carmo}
Taneesh Gupta, Shivam Shandilya, Xuchao Zhang, Rahul Madhavan, Supriyo Ghosh,
  Chetan Bansal, Huaxiu Yao, and Saravan Rajmohan.
\newblock \href{https://aclanthology.org/2025.findings-acl.114/}{{CARMO}: Dynamic criteria generation for context aware reward
  modelling.}
\newblock In Wanxiang Che, Joyce Nabende, Ekaterina Shutova, and Mohammad~Taher
  Pilehvar, editors, {\em Findings of the {A}ssociation for {C}omputational
  {L}inguistics: {ACL} 2025}, pages 2202--2261, Vienna, Austria, July 2025.
  Association for Computational Linguistics.

\bibitem[HER{\etalchar{+}}24]{hashemi2024rubric}
Helia Hashemi, Jason Eisner, Corby Rosset, Benjamin Van~Durme, and Chris
  Kedzie.
\newblock \href{https://aclanthology.org/2024.acl-long.745/}{{LLM}-rubric: A multidimensional, calibrated approach to automated
  evaluation of natural language texts.}
\newblock In Lun-Wei Ku, Andre Martins, and Vivek Srikumar, editors, {\em
  Proceedings of the 62nd Annual Meeting of the {A}ssociation for
  {C}omputational {L}inguistics (Volume 1: Long Papers)}, pages 13806--13834,
  Bangkok, Thailand, August 2024. Association for Computational Linguistics.

\bibitem[KL17]{koh2017influence}
Pang~Wei Koh and Percy Liang.
\newblock \href{https://proceedings.mlr.press/v70/koh17a.html}{Understanding black-box predictions via influence functions.}
\newblock In Doina Precup and Yee~Whye Teh, editors, {\em Proceedings of the
  34th International Conference on Machine Learning}, volume~70 of {\em
  Proceedings of Machine Learning Research}, pages 1885--1894. PMLR, 06--11 Aug
  2017.

\bibitem[KSD{\etalchar{+}}26]{kola2026saja}
Sneha Kola, Pankaj~Kumar Sharma, Soumyadeep Dey, Bamdev Mishra, and Mayur
  Datar.
\newblock \href{https://aclanthology.org/2026.acl-industry.45/}{{SAJA}: A simple approach to judge alignment for {LLM}-as-a-judge.}
\newblock In Yunyao Li, Georg Rehm, and Mei Tu, editors, {\em Proceedings of
  the 64th Annual Meeting of the {A}ssociation for {C}omputational
  {L}inguistics (Volume 6: Industry Track)}, pages 646--664, San Diego,
  California, USA, July 2026. Association for Computational Linguistics.

\bibitem[KSL{\etalchar{+}}24]{kim2024prometheus2}
Seungone Kim, Juyoung Suk, Shayne Longpre, Bill~Yuchen Lin, Jamin Shin, Sean
  Welleck, Graham Neubig, Moontae Lee, Kyungjae Lee, and Minjoon Seo.
\newblock \href{https://aclanthology.org/2024.emnlp-main.248/}{{Prometheus 2}: An open source language model specialized in
  evaluating other language models.}
\newblock In Yaser Al-Onaizan, Mohit Bansal, and Yun-Nung Chen, editors, {\em
  Proceedings of the 2024 Conference on Empirical Methods in Natural Language
  Processing}, pages 4334--4353, Miami, Florida, USA, November 2024.
  Association for Computational Linguistics.

\bibitem[LCA{\etalchar{+}}25]{li2025calibraeval}
Haitao Li, Junjie Chen, Qingyao Ai, Zhumin Chu, Yujia Zhou, Qian Dong, and
  Yiqun Liu.
\newblock \href{https://aclanthology.org/2025.acl-long.808/}{{C}alibra{E}val: Calibrating prediction distribution to mitigate
  selection bias in {LLM}s-as-judges.}
\newblock In Wanxiang Che, Joyce Nabende, Ekaterina Shutova, and Mohammad~Taher
  Pilehvar, editors, {\em Proceedings of the 63rd Annual Meeting of the
  {A}ssociation for {C}omputational {L}inguistics (Volume 1: Long Papers)},
  pages 16537--16552, Vienna, Austria, July 2025. Association for Computational
  Linguistics.

\bibitem[LIX{\etalchar{+}}23]{liu2023geval}
Yang Liu, Dan Iter, Yichong Xu, Shuohang Wang, Ruochen Xu, and Chenguang Zhu.
\newblock \href{https://aclanthology.org/2023.emnlp-main.153/}{{G}-eval: {NLG} evaluation using {GPT}-4 with better human alignment.}
\newblock In Houda Bouamor, Juan Pino, and Kalika Bali, editors, {\em
  Proceedings of the 2023 Conference on Empirical Methods in Natural Language
  Processing}, pages 2511--2522, Singapore, December 2023. Association for
  Computational Linguistics.

\bibitem[LPM{\etalchar{+}}24]{lee2024rlaif}
Harrison Lee, Samrat Phatale, Hassan Mansoor, Thomas Mesnard, Johan Ferret,
  Kellie~Ren Lu, Colton Bishop, Ethan Hall, Victor Carbune, Abhinav Rastogi,
  and Sushant Prakash.
\newblock \href{https://proceedings.mlr.press/v235/lee24t.html}{{RLAIF} vs. {RLHF}: Scaling reinforcement learning from human
  feedback with {AI} feedback.}
\newblock In Ruslan Salakhutdinov, Zico Kolter, Katherine Heller, Adrian
  Weller, Nuria Oliver, Jonathan Scarlett, and Felix Berkenkamp, editors, {\em
  Proceedings of the 41st International Conference on Machine Learning}, volume
  235 of {\em Proceedings of Machine Learning Research}, pages 26874--26901.
  PMLR, 21--27 Jul 2024.

\bibitem[LXY{\etalchar{+}}26]{liu2026openrubrics}
Tianci Liu, Ran Xu, Tony Yu, Ilgee Hong, Carl Yang, Tuo Zhao, and Haoyu Wang.
\newblock \href{https://aclanthology.org/2026.acl-long.791/}{{O}pen{R}ubrics: Towards scalable synthetic rubric generation for
  reward modeling and {LLM} alignment.}
\newblock In Maria Liakata, Viviane~P. Moreira, Jiajun Zhang, and David
  Jurgens, editors, {\em Proceedings of the 64th Annual Meeting of the
  {A}ssociation for {C}omputational {L}inguistics (Volume 1: Long Papers)},
  pages 17417--17437, San Diego, California, United States, July 2026.
  Association for Computational Linguistics.

\bibitem[ME20]{mehri2020usr}
Shikib Mehri and Maxine Eskenazi.
\newblock \href{https://aclanthology.org/2020.acl-main.64/}{{USR}: An unsupervised and reference free evaluation metric for
  dialog generation.}
\newblock In Dan Jurafsky, Joyce Chai, Natalie Schluter, and Joel Tetreault,
  editors, {\em Proceedings of the 58th Annual Meeting of the Association for
  Computational Linguistics}, pages 681--707, Online, July 2020. Association
  for Computational Linguistics.

\bibitem[MHBS26]{mukherjee2026geometry}
Sourabrata Mukherjee, Hamna Hamna, Kalika Bali, and Sunayana Sitaram.
\newblock \href{https://arxiv.org/abs/2606.03043}{The geometry of {LLM}-as-judge: Why inter-{LLM} consensus is not
  human alignment, 2026.}
\newblock Accepted to the 2026 Conference on Empirical Methods in Natural
  Language Processing ({EMNLP}).
\newblock arXiv:2606.03043.

\bibitem[PBF24]{panickssery2024selfpreference}
Arjun Panickssery, Samuel~R. Bowman, and Shi Feng.
\newblock \href{https://proceedings.neurips.cc/paper_files/paper/2024/file/7f1f0218e45f5414c79c0679633e47bc-Paper-Conference.pdf}{{LLM} evaluators recognize and favor their own generations.}
\newblock In A.~Globerson, L.~Mackey, D.~Belgrave, A.~Fan, U.~Paquet,
  J.~Tomczak, and C.~Zhang, editors, {\em Advances in Neural Information
  Processing Systems}, volume~37, pages 68772--68802. Curran Associates, Inc.,
  2024.

\bibitem[RCB26]{rao2026autorubric}
Delip Rao and Chris Callison-Burch.
\newblock \href{https://arxiv.org/abs/2603.00077}{{Autorubric}: A unifying framework for rubric-based {LLM} evaluation
  on non-verifiable tasks.}
\newblock In {\em The Third Annual Conference on Language Modeling ({COLM})},
  2026.
\newblock Camera-ready version: arXiv:2603.00077.

\bibitem[RSG16]{ribeiro2016trust}
Marco~Tulio Ribeiro, Sameer Singh, and Carlos Guestrin.
\newblock \href{http://dx.doi.org/10.1145/2939672.2939778}{{``Why Should I Trust You?''}: Explaining the predictions of any
  classifier.}
\newblock In {\em Proceedings of the 22nd ACM SIGKDD International Conference
  on Knowledge Discovery and Data Mining}, KDD '16, pages 1135--1144. ACM,
  August 2016.

\bibitem[SAA26]{siro2026learning}
Clemencia Siro, Pourya Aliannejadi, and Mohammad Aliannejadi.
\newblock \href{https://aclanthology.org/2026.findings-eacl.335/}{Learning to judge: {LLM}s designing and applying evaluation rubrics.}
\newblock In Vera Demberg, Kentaro Inui, and Llu{\'i}s Marquez, editors, {\em
  Findings of the {A}ssociation for {C}omputational {L}inguistics: {EACL}
  2026}, pages 6371--6389, Rabat, Morocco, March 2026. Association for
  Computational Linguistics.

\bibitem[SML{\etalchar{+}}25]{shi2025judges}
Lin Shi, Chiyu Ma, Wenhua Liang, Xingjian Diao, Weicheng Ma, and Soroush
  Vosoughi.
\newblock \href{https://aclanthology.org/2025.ijcnlp-long.18/}{Judging the judges: A systematic study of position bias in
  {LLM}-as-a-judge.}
\newblock In Kentaro Inui, Sakriani Sakti, Haofen Wang, Derek~F. Wong, Pushpak
  Bhattacharyya, Biplab Banerjee, Asif Ekbal, Tanmoy Chakraborty, and
  Dhirendra~Pratap Singh, editors, {\em Proceedings of the 14th International
  Joint Conference on Natural Language Processing and the 4th Conference of the
  {A}sia-Pacific Chapter of the {A}ssociation for {C}omputational
  {L}inguistics}, pages 292--314, Mumbai, India, December 2025. The Asian
  Federation of Natural Language Processing and The Association for
  Computational Linguistics.

\bibitem[TZM{\etalchar{+}}25]{tan2025judgebench}
Sijun Tan, Siyuan Zhuang, Kyle Montgomery, William~Y. Tang, Alejandro Cuadron,
  Chenguang Wang, Raluca~Ada Popa, and Ion Stoica.
\newblock \href{https://proceedings.iclr.cc/paper_files/paper/2025/file/9e720fce64f91114c49cfd640d821da3-Paper-Conference.pdf}{{JudgeBench}: A benchmark for evaluating {LLM}-based judges.}
\newblock In Y.~Yue, A.~Garg, N.~Peng, F.~Sha, and R.~Yu, editors, {\em
  International Conference on Learning Representations}, volume 2025, pages
  63277--63303, 2025.

\bibitem[WDD{\etalchar{+}}24]{wang2024helpsteer2}
Zhilin Wang, Yi~Dong, Olivier Delalleau, Jiaqi Zeng, Gerald Shen, Daniel Egert,
  Jimmy~J. Zhang, Makesh~Narsimhan Sreedhar, and Oleksii Kuchaiev.
\newblock \href{https://proceedings.neurips.cc/paper_files/paper/2024/file/02fd91a387a6a5a5751e81b58a75af90-Paper-Datasets_and_Benchmarks_Track.pdf}{{HelpSteer 2}: Open-source dataset for training top-performing reward
  models.}
\newblock In A.~Globerson, L.~Mackey, D.~Belgrave, A.~Fan, U.~Paquet,
  J.~Tomczak, and C.~Zhang, editors, {\em Advances in Neural Information
  Processing Systems}, volume~37, pages 1474--1501. Curran Associates, Inc.,
  2024.

\bibitem[WLC{\etalchar{+}}24]{wang2024fair}
Peiyi Wang, Lei Li, Liang Chen, Zefan Cai, Dawei Zhu, Binghuai Lin, Yunbo Cao,
  Lingpeng Kong, Qi~Liu, Tianyu Liu, and Zhifang Sui.
\newblock \href{https://aclanthology.org/2024.acl-long.511/}{Large language models are not fair evaluators.}
\newblock In Lun-Wei Ku, Andre Martins, and Vivek Srikumar, editors, {\em
  Proceedings of the 62nd Annual Meeting of the {A}ssociation for
  {C}omputational {L}inguistics (Volume 1: Long Papers)}, pages 9440--9450,
  Bangkok, Thailand, August 2024. Association for Computational Linguistics.

\bibitem[YWH{\etalchar{+}}25]{ye2025justice}
Jiayi Ye, Yanbo Wang, Yue Huang, Dongping Chen, Qihui Zhang, Nuno Moniz, Tian
  Gao, Werner Geyer, Chao Huang, Pin-Yu Chen, Nitesh~V. Chawla, and Xiangliang
  Zhang.
\newblock \href{https://proceedings.iclr.cc/paper_files/paper/2025/file/fdca08d371e4b6c031397909e20043bd-Paper-Conference.pdf}{Justice or prejudice? quantifying biases in {LLM}-as-a-judge.}
\newblock In Y.~Yue, A.~Garg, N.~Peng, F.~Sha, and R.~Yu, editors, {\em
  International Conference on Learning Representations}, volume 2025, pages
  102351--102390, 2025.

\bibitem[ZCS{\etalchar{+}}23]{zheng2023judging}
Lianmin Zheng, Wei-Lin Chiang, Ying Sheng, Siyuan Zhuang, Zhanghao Wu, Yonghao
  Zhuang, Zi~Lin, Zhuohan Li, Dacheng Li, Eric~P. Xing, Hao Zhang, Joseph~E.
  Gonzalez, and Ion Stoica.
\newblock \href{https://proceedings.neurips.cc/paper_files/paper/2023/file/91f18a1287b398d378ef22505bf41832-Paper-Datasets_and_Benchmarks.pdf}{Judging {LLM}-as-a-judge with {MT}-bench and {Chatbot Arena}.}
\newblock In A.~Oh, T.~Naumann, A.~Globerson, K.~Saenko, M.~Hardt, and
  S.~Levine, editors, {\em Advances in Neural Information Processing Systems},
  volume~36, pages 46595--46623. Curran Associates, Inc., 2023.

\bibitem[ZSZ{\etalchar{+}}26]{zufle2026calibrating}
Maike Z{\"u}fle, Patr{\'i}cia Schmidtov{\'a}, Vil{\'e}m Zouhar, Shree~Harsha
  Bokkahalli~Satish, Erica Cooper, Shobhit Banga, Vaibhav Nalawade, Manmeet
  Kaur, Jan Niehues, Markus M{\"u}ller, and Ond{\v{r}}ej Klejch.
\newblock \href{https://arxiv.org/abs/2609.29431}{Calibrating {LLM} judges for human and {AI} conversations, 2026.}
\newblock arXiv:2609.29431.

\bibitem[ZZW{\etalchar{+}}26]{zhou2026rubricbench}
Junyi Zhou, Qiyuan Zhang, Yufei Wang, Fuyuan Lyu, Yidong Ming, Can Xu, Qingfeng
  Sun, Kai Zheng, Peng Kang, Xue Liu, and Chen Ma.
\newblock \href{https://aclanthology.org/2026.acl-long.1439/}{{R}ubric{B}ench: Aligning model-generated rubrics with human
  standards.}
\newblock In Maria Liakata, Viviane~P. Moreira, Jiajun Zhang, and David
  Jurgens, editors, {\em Proceedings of the 64th Annual Meeting of the
  {A}ssociation for {C}omputational {L}inguistics (Volume 1: Long Papers)},
  pages 31179--31200, San Diego, California, United States, July 2026.
  Association for Computational Linguistics.

\end{thebibliography}
\end{document}